\documentclass{article}
\usepackage{arxiv}

\usepackage{amsmath,amsfonts}
\usepackage{algorithmic}
\usepackage{algorithm}
\usepackage{array}
\usepackage{textcomp}
\usepackage{stfloats}
\usepackage{url}
\usepackage{verbatim}
\usepackage{graphicx}
\usepackage{cite,comment}
\newcounter{daggerfootnote}

\usepackage{graphicx}
\usepackage{tikz}
\usepackage{comment}
\usepackage{amsmath,amssymb} % define this before the line numbering.
\usepackage{color}

\usepackage{amssymb}% http://ctan.org/pkg/amssymb
\usepackage{pifont}% http://ctan.org/pkg/pifont
\newcommand{\cmark}{\ding{51}}
\newcommand{\xmark}{{\color{red} \ding{55}}}%

\usepackage[accsupp]{axessibility}  % Improves PDF readability for those with disabilities.

\usepackage{subfig}
\usepackage{footmisc}
\usepackage{stfloats}
\usepackage{multicol}
\usepackage{marvosym}
\usepackage{multirow}
\usepackage{times}
\usepackage{epsfig}
\usepackage{bm}
\usepackage{comment}
\usepackage{tablefootnote}

\usepackage[utf8]{inputenc} % allow utf-8 input
\usepackage[T1]{fontenc}    % use 8-bit T1 fonts
\usepackage{hyperref}       % hyperlinks
\usepackage{url}            % simple URL typesetting
\usepackage{booktabs}       % professional-quality tables
\usepackage{amsfonts}       % blackboard math symbols
\usepackage{nicefrac}       % compact symbols for 1/2, etc.
\usepackage{microtype}      % microtypography
\usepackage{xcolor}         % colors

\title{Cell Tracking-by-detection using Elliptical Bounding Boxes}

\author{ \href{https://orcid.org/0000-0001-6611-5327}{\includegraphics[scale=0.06]{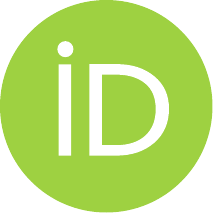}\hspace{1mm}Lucas N.~Kirsten}\\
	Institute of Informatics\\
	Universidade Federal do Rio Grande do Sul\\
	Porto Alegre, RS (Brazil) \\
	\texttt{lnkirsten@inf.ufrgs.br} \\
	\And
	\href{https://orcid.org/0000-0002-4711-5783}{\includegraphics[scale=0.06]{orcid.pdf}\hspace{1mm}Cláudio R.~Jung} \\
	Institute of Informatics\\
	Universidade Federal do Rio Grande do Sul\\
	Porto Alegre, RS (Brazil) \\
	\texttt{crjung@inf.ufrgs.bru} \\
}

\renewcommand{\shorttitle}{Cell Tracking-by-detection using Elliptical Bounding Boxes}

\hypersetup{
pdftitle={Cell Tracking-by-detection using Elliptical Bounding Boxes},
pdfauthor={Lucas N.~Kirten, Cláudio R.~Jung},
pdfkeywords={Cell tracking, Cell detection, Oriented object detection},
}

\begin{document}
\maketitle

\begin{abstract}
Cell detection and tracking are paramount for bio-analysis. Recent approaches rely on the tracking-by-model evolution paradigm, which usually consists of training end-to-end deep learning models to detect and track the cells on the frames with promising results. However, such methods require extensive amounts of annotated data, which is time-consuming to obtain and often requires specialized annotators. This work proposes a new approach based on the classical tracking-by-detection paradigm that alleviates the requirement of annotated data. More precisely, it approximates the cell shapes as oriented ellipses and then uses generic-purpose oriented object detectors to identify the cells in each frame. We then rely on a global data association algorithm that explores temporal cell similarity using probability distance metrics, considering that the ellipses relate to two-dimensional Gaussian distributions. Our results show that our method can achieve detection and tracking results competitively with state-of-the-art techniques that require considerably more extensive data annotation. Our code is available at: \url{https://github.com/LucasKirsten/Deep-Cell-Tracking-EBB}.
\end{abstract}


\maketitle

\section{Introduction}
Detection and tracking of living cells in microscopy images is a crucial task required in many biomedical applications, such as cell growth, migration, invasion, morphological changes, and changes in the localization of molecules within cells~\cite{syed2008detection,leite2015computational,di2019learning,gradeci2020single}. The sheer amount of data produced by high-throughput microscopy imaging imposes an analytical challenge for science researchers, which can only be overcome with the appropriate computational tools. 

As with several other computer vision tasks, the state-of-the-art (SOTA) for cell detection and tracking is based on deep learning approaches~\cite{hayashida2022consistent,emami2021computerized}. These techniques typically require manual cell annotations for training and evaluating the models, and the annotation format has a significant impact on both the time devoted to image labeling and the complexity of the network itself. The most traditional object representation refers to using horizontal bounding boxes (HBBs, a.k.a. BBs) to detect objects in a scene. Despite being very simple to annotate, this representation is not adequate when dealing with oriented elongated objects, since the HBB may contain large portions of the background or other objects in clutter scenarios. On the other hand, segmenting each object provides a fine-grained representation of the shape, but it is a tedious and time-consuming task. Moreover, applications that use multiple cell lineages from different sources (e.g., microscope, cell type) may require several rounds of labeling data and retraining the models~\cite{ulman2017objective}. In these cases, a fast and efficient method for quickly labeling the data is crucial for the application continuity, since it is usually the most time-consuming step.

\begin{figure}[]
    \centering
    \includegraphics[width=.4\textwidth]{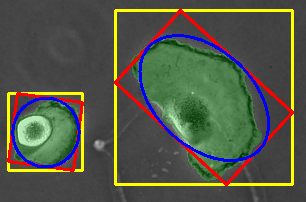}
    \caption{Comparison of different types of annotations. In green are the full segmentation mask, in yellow the HBB representation, in red the OBB representation, and in blue the EBB representation.}
    \label{fig:comparison_annotations}
\end{figure}

In the context of cell detection and tracking, knowing the complete shape representation might not be needed, while it can impose a real challenge in cases where the cell contour is highly uncertain (e.g., when there is low contrast between the cells and the background). Furthermore,  methods used for individually segmenting the cell masks usually require more complex and computationally expensive algorithms, since they are typically developed in a two-step manner (either detecting and then segmenting~\cite{cpn,maskrcnn}, or segmenting and then splitting the masks~\cite{bise,unet,gcme,unets,drl}). Oriented bounding boxes (OBBs, a.k.a. rotated bounding boxes) are an intermediate representation between segmentation masks and HBBs with a good compromise between simplicity and completeness. However, the presence of roughly circular cells imposes angular ambiguity on its representation, since its OBB representation will be a square rotated at any angle (see the left cell on Figure~\ref{fig:comparison_annotations}). 

In this work, we advocate using elliptical bounding boxes (EBBs), which can be directly derived from OBBs and can capture oriented cells while mitigating the angular ambiguity for circular objects. Fig.~\ref{fig:comparison_annotations} shows a comparison of HBBs, OBBs, segmentation masks, and EBBs for two different cells: the left one is roughly circular, and right one is oriented. The EBB representation, shown in blue, presents a good fit in both examples. As an additional advantage, the orientation/shape of detected cells represented as EBBs can be explored in tracking-by-detection approaches to provide a better spatio-temporal association when time-lapse sequences are used, while weakly-supervised segmentation methods such as~\cite{kulharia2020box2seg} can be coupled to the detected OBBs/EBBs to obtain a more detailed representation of the cell shape.

We propose a cell tracking-by-detection method that uses a deep learning model to detect the cells as OBBs, then convert them to EBBs to fit the cell shapes better. For the tracking part, we based our solution on the work of Bise et~al.~\cite{bise}, which describes an unsupervised (i.e., no tracking label is necessary) long-term global data association algorithm. We adapted their algorithm to rely only on the detection information provided by the object detector model (e.g., position, confidence score). It is important to note that our method only requires OBB cell annotations for isolated frames, and no tracking annotation involving temporal sequences is needed. 

\section{Related work}

\subsection{Cell detection and segmentation}
There are several approaches for cell counting, detection, and segmentation, and the best results have been achieved by deep learning methods~\cite{xing2017deep,moen2019deep}. 
These methods vary considerably regarding the underlying structure of the network and also on the degree of supervision required to label training data. For example, detecting just the cell nucleus requires one pixel-per-cell as supervision; detecting the cell boundaries as an HBB requires two points (top-left and bottom-right), while OBBs require an additional parameter related to the orientation; finally, segmentation requires the identification of all pixels belonging to a cell, which is very time-consuming. There are also some intermediate shape representations such as the \textit{star-convex polygons}~\cite{schmidt2018cell}.

The U-Net presented in~\cite{unet} has become a popular segmentation approach in biomedical applications when segmentation-level information is required. It is based on an encoder-decoder U-shaped architecture with skip connections, and focuses mostly on \textit{semantic segmentation} tasks (i.e., segmenting all objects for each class altogether). Although the connected components produced by U-Net can also be used for \emph{instance segmentation}, dense scenarios or situations with strongly overlapping cells are challenging. For instance segmentation, most approaches produce an embedding vector for each image pixel in such a way that similar vectors should relate to the same instance (object)~\cite{newell2017associative}, and have been explored in the context of cell segmentation by Payer and colleagues~\cite{payer2018instance,payer2019segmenting} by using a cosine loss for estimating local embedding distances. To mitigate the cost of clustering pixels embedding (that is required to obtain the final instance segmentation), Zhao and colleagues~\cite{zhao2021faster} proposed a fast Mean-Shift algorithm that works on GPUs. The \textit{panoptic segmentation} task combines both category- and instance-level information into a single framework, and has been applied to biomedical imaging by Liu and colleagues~\cite{liu2021panoptic}.

Although solutions that can return the full segmentation mask of cells have the clear advantage of providing a complete shape representation, most segmentation approaches are fully supervised, requiring time-consuming per-cell annotations. In order to overcome this problem, weakly supervised approaches explore partial annotations such as bounding boxes~\cite{zhao2018deep}, center-points~\cite{zhao2020weakly} or user-defined scribbles~\cite{oh2022scribble} for the segmentation task. Nevertheless, many applications (e.g., tissue engineering~\cite{lu2021biofabrication}) do not require the full description of the cell shape, but rather a position, size, and orientation descriptor are enough. This alleviates the annotation process, making it faster, less tedious, and more scalable.

As noted in~\cite{schmidt2018cell}, a popular approach for cell detection in microscopy images is to use generic-purpose object detectors based on HBBs, such as SSD~\cite{liu2016ssd} or YOLO~\cite{redmon2016you}. Despite the constant evolution of object detectors~\cite{carion2020end,tan2020efficientdet,Wang_2021_CVPR,yang2022focal}, the representation of cells as HBBs presents limitations in denser scenarios, particularly when oriented and elongated cells are present as aforementioned. Intermediate representations between HBBs and full segmentation masks have also been explored for cell detection. For example, Schmidt~et~al.~\cite{schmidt2018cell} presented a polygonal shape representation based on radial sweeps with equidistant angles. A similar approach using splines instead of polygonal representations was presented in~\cite{mandal2021splinedist}, obtaining smoother cell boundaries. 

In this work, we advocate using EBB representations for cell detection, which is a natural extension of the OBB representation, but usually provides a better fit to the cell shape. In particular, roughly circular shapes induce a naturally ambiguous angular representation when OBBs are used since any rotated square fits equally to a circle. For these cases, the EBB would reduce to a circle, mitigating the angular problem. Finally, it is worth mentioning that the output of any OBB detector can be mapped to an EBB, and OBB detectors have demonstrated exciting results in the past years~\cite{r2cnn,yang2021r3det,dota,kld}. An OBB annotation is almost as simple as an HBB, with a considerable gain in shape representation, particularly regarding applications that involve elongated cells.

\subsection{Cell tracking}

Firstly, it is important to emphasize that ``cell tracking'' presents a broad spectrum of challenges, since each cell type and application might require a tailored solution. As noted in \cite{ulman2017objective}, ``there is no simple way to point out the right algorithm for a given dataset'', hinting that finding an algorithm capable of working for a broad spectrum of cell lineages is a huge challenge. Nevertheless, the ISBI Challenge~\cite{isbi} has provided several different datasets for benchmarking cell tracking algorithms, which usually are elaborated to work on more than one cell type and lineage. 

Recent cell tracking algorithms can be broadly divided into two categories~\cite{epflheid,blob,drl}: (i) tracking by model evolution and (ii) tracking-by-detection. In tracking by model evolution methods (a.k.a. end-to-end tracking systems), detection and tracking  are solved simultaneously. In this context, Payer~et~al.~\cite{payer2018instance,payer2019segmenting} introduced temporal information for spatio-temporal learning of the embeddings. They explored a cosine loss for estimating local embedding distances, and used a convolutional Gated Recursive Unit (ConvGRU) to learn temporal relationships. Nishimura~et~al.~\cite{nishimura2020weakly} presented a cell tracking approach that works with weak annotation (cell centers in successive frames) by exploring a co-detection CNN. More recently, Hayashida~et~al.~\cite{hayashida2022consistent} proposed a complete pipeline that uses spatial-temporal context in multiple frames and long-term motion estimation with an objective level warping loss that addresses the problem of detecting and tracking highly dense cell images. Although these methods are capable of achieving high performance, it is important to emphasize that they all require full annotations for both the detection/segmentation and tracking steps, which might be a strong limitation.

The tracking-by-detection paradigm consists of two stages: cell detection and cell association. When segmentation masks are required, the detection stage can either aid a segmentation step to extract the masks of each detected cell \cite{cpn}, or it can directly infer the segmentation masks and then split those wrongly joined cells using some algorithm such as the watershed \cite{kth,boukari2018automated,epflheid,blob,unet,gcme}. The methods for connecting the cells in subsequent frames usually rely on graphs and linear integer programming~\cite{kth,boukari2018automated,epflheid,blob,cpn} by defining the costs of the cell events (e.g., movement, mitoses, and apoptosis). 
%Other methods used to connect the cells relate to active tracks~\cite{unet} or growing of the segmentation masks on subsequent frames~\cite{gcme}. 
Other strategies include the use of multi-Bernoulli random finite sets~\cite{xu2019automated} or joint particle filters based on Markov random field to model the dependency of the target movements~\cite{hirose2017spf}.
Akram~et~al.~\cite{cpn} uses a deep learning model that first detects the cells using the HBB representation and then feeds these detections to a segmentation model to further retrieve the segmentation masks of individual cells. For associating the detection, they use random forests to estimate the costs of the event graph. More recently, Wang~et~al.~\cite{drl} proposed a method that first segments the images using a U-Net~\cite{unet} deep learning model, and then uses deep reinforcement learning to associate the detected targets between frames.

This work proposes a tracking-by-detection approach that solely uses spatial information and the detection scores from an object detector to determine the associations between cells. It alleviates the annotation requirements to use only annotated cells as OBBs, and also eliminates the necessity of using other features from the detections (e.g., image histograms) to compute the associations.

\section{The proposed tracking-by-detection method}\label{sec:method}

\begin{figure*}[]
    \centering
    \includegraphics[width=0.95\textwidth]{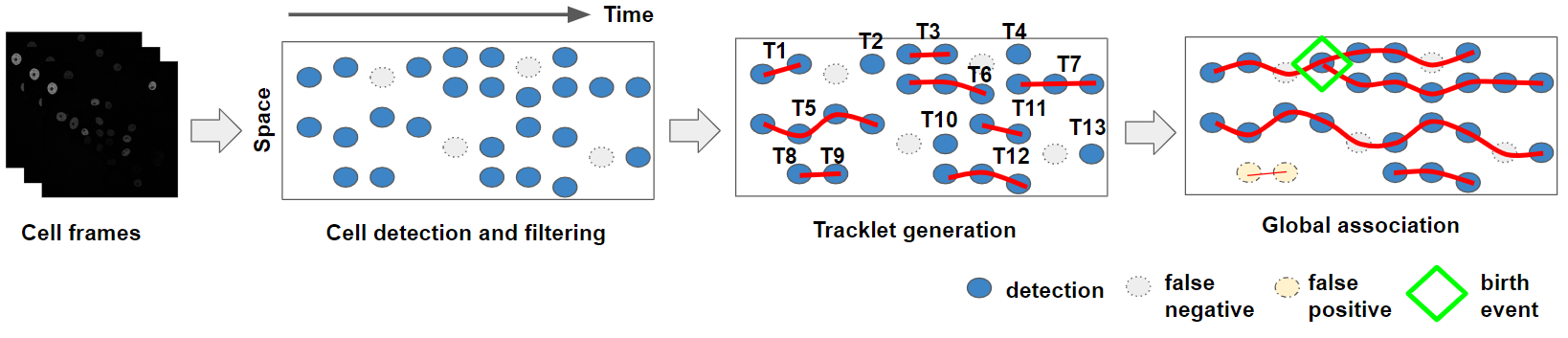}
    \caption[Overview of the proposed pipeline for cell-tracking-by-detection.]{Overview of the proposed pipeline for cell-tracking-by-detection. First, we detect the cells as OBBs and then convert them to the EBB representation. Next, we join the cells with high overlap between two adjecent frames. Finally, a global data association algorithm is used to identify all the cell events (i.e., movement, mitoses and apoptosis), while filling gaps generated by false negative detections and removing false positives ones, in order to produce the final tracklets.}
    \label{fig:complete_pipeline}
\end{figure*}

Our approach follows the typical pipeline of a tracking-by-detection method. First, an object detector is used to identify the cells in each frame using OBB representations, which are then converted to elliptical representations (EBBs). For tracking, we initially generate short tracklets by joining cells in subsequent frames with an objective function that jointly explores the shape and distance of cells. More precisely, we map the EBB representation to a two-dimensional Gaussian distribution and explore the Helinger distance, which directly correlates to the IoU metric~\cite{probiou}. Finally, a global data association method based on the work of \cite{bise} associates the tracklets to obtain the final cell trajectories and lineage trees. Figure~\ref{fig:complete_pipeline} shows an overview of our complete pipeline for cell tracking-by-detection, and the steps are detailed next.

\subsection{Cell detection}

The first step of our method relies on identifying the cells for each frame individually. 
We propose to use off-the-shelf OBB object detectors trained with cell images and then convert the output to elliptical bounding boxes (EBBs). For an OBB with center $(x,y)$, width $W$, height $H$, and orientation $\theta$, we generate an ellipse with the same center and orientation, with semi-axes $a=W/2$ and $b=H/2$. If the OBB is clearly oriented (i.e., $W >\!\!> H$ or $W <\!\!< H$), the EBB will preserve the orientation of the OBB. On the other hand, if the OBB is roughly square (i.e., $H\approx W$), the produced EBB will be roughly circular. In the case of a perfect square, the EBB simplifies to a single circle regardless of the orientation of the OBB, which mitigates the orientation ambiguity (recall the example shown in Fig.~\ref{fig:comparison_annotations}).

\subsection{Detection filtering and suppression}
\label{sec:detection_filtering}

In a typical deep object detector, only candidate detections with scores larger than a pre-defined threshold $\tau_s$ are retrieved. Still, we usually have several overlapping candidates related to the same object, and Non-Maximum Suppression (NMS) is then used to retrieve only the candidate with the highest score. In this step, it is crucial to define a geometrical similarity measure between the detections for quantifying their ``overlap'' degree. 

The Intersection-over-Union (IoU) is the \textit{de facto} standard metric for computing the overlap in HBB or OBB detectors. However, computing the IoU for OBBs is not trivial due to the several possibilities for two intersecting OBBs~\cite{chen2020piou}. Furthermore, the IoU is unreliable for OBB detections related to circular cells, since angular discrepancies might artificially degrade the IoU~\cite{Murrugarra-Llerena_2022_CVPR}. Using the IoU with EBBs mitigates the latter problem, because the ellipse reduces to a circle. However, computing the intersection using EBBs is even more complex than using OBBs since it involves the overlap of two ellipses. In this work, we propose an alternative similarity metric based on fuzzy object representations.

In \cite{gwd,kld} and \cite{probiou}, the core idea is to use 2D Gaussian distributions (denoted as GBBs -- Gaussian Bounding Boxes) as intermediate representations for oriented objects, and train OBB detectors using loss functions based on similarity metrics between distributions. Here, we explore their developed fuzzy representation to compute the distance/similarity, as explained next. Following \cite{probiou}, an OBB with center $\bm{\mu} = (x_c, y_c)^T$, width $W$, height $H$ and angle $\theta$ is mapped to a GBB described by the mean vector $\bm{\mu}$ (which is the OBB center) and a covariance matrix \begin{equation}
\small
    \Sigma =
    \begin{bmatrix}a & c \\
    c  & b\end{bmatrix}
    = 
    \begin{bmatrix}\frac{W^2}{12} \cos^{2}\theta + \frac{H^2}{12} \sin^{2}\theta & \frac{1}{2}\left(\frac{W^2}{12} - \frac{H^2}{12}\right) \sin2 \theta \\
    \frac{1}{2}\left(\frac{W^2}{12} - \frac{H^2}{12}\right) \sin2 \theta  & \frac{W^2}{12} \sin^{2}\theta  + \frac{H^2}{12} \cos^{2}\theta\end{bmatrix}.
\end{equation}
Note that square OBBs, for which $H=W$, generate a diagonal covariance matrix that does not involve the angular parameter $\theta$. Hence, squares that differ only by angle are mapped to the exact same GBB.

In \cite{gwd,kld} and \cite{probiou}, the Gaussian distributions are used to train OBB object detectors as their loss function. However, the metrics proposed by \cite{gwd,kld} do not hold the mathematical properties of a similarity measure, which is the goal for comparing two detection.
Therefore, we define a similarity metric based on the Hellinger Distance, as done by \cite{probiou}. Let us consider that $p\sim\mathcal{N}(\bm{\mu}_1, \Sigma_1)$ and $q\sim\mathcal{N}(\bm{\mu}_2, \Sigma_2)$ are Gaussian distributions with 
\begin{equation}
    \bm{\mu}_1 = \begin{pmatrix}
    x_1 \\ y_1
    \end{pmatrix},~
    \Sigma_1 = \begin{bmatrix}
    a_1 & c_1\\
    c_1 & b_1 
    \end{bmatrix},~
    \bm{\mu}_2  = \begin{pmatrix}
    x_2 \\ y_2
    \end{pmatrix},~
    \Sigma_2 = \begin{bmatrix}
    a_2 & c_2\\
    c_2 & b_2 
    \end{bmatrix}.
\end{equation}
The Bhattacharyya Distance~\cite{bhattacharyya1946measure} between distributions $p$ and $q$ is given by 
\begin{align}
\label{eq:bhatta}
B_D(p,q) &= \frac{1}{8}\bm{\mu_{12}}^T\Sigma^{-1}\bm{\mu_{12}} +   \frac{1}{2}\ln \left( \frac{\det\Sigma_{12}}{\sqrt{\det\Sigma_1 \det\Sigma_2}}  \right), \\\nonumber
\bm{\mu_{12}} &= \bm{\mu}_1 - \bm{\mu}_2,~ \Sigma_{12} = \frac{1}{2}\left(\Sigma_1 + \Sigma_2\right),
\end{align}
and the Hellinger distance~\cite{hellinger1909neue} between $p$ and $q$ is then
\begin{equation}
    H_D(p,q) =\sqrt{1-e^{-B_D(p,q)}}.
    \label{eq:helinger}
\end{equation}

\begin{figure}[]
    \centering
    \subfloat[][Frame with detections before filtering and suppression.]{\includegraphics[width = 0.4\textwidth]{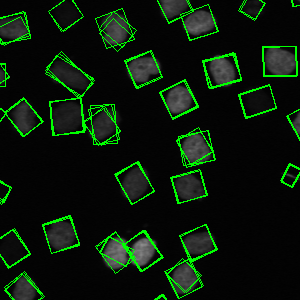}}~~
    \subfloat[][Frame with detections after filtering and suppression.]{\includegraphics[width = 0.4\textwidth]{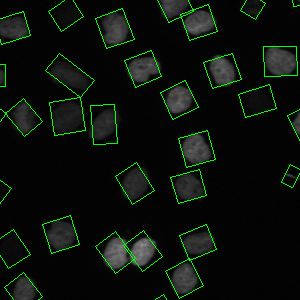}}
    \caption[Illustration of detection filtering and NMS using the Helinger Distance on a frame.]{Illustration of detection filtering and NMS using the Hellinger Distance on a frame. Observe that in (a) we have many duplicates and false positives detection, while in (b) they are suppressed and eliminated, resulting in only one detection per cell.}
    \label{fig:hd_nms}
\end{figure}

Although both $B_D$ and $H_D$ are named ``distances'', only $H_D$ satisfies the mathematical properties for a metric~\cite{kailath1967divergence}. Moreover, we can see that $0 \leq H_D(p,q) \leq 1$ (with $0$ being the maximum similarity), meaning that $H_D$ provides a normalized distance metric. In this work, we explore $H_D$ to find overlapping EBBs to suppress non-maximum detections. Finally, given a detection $p$ with the highest confidence score, we suppress all detections $q\neq p$ for which  $H_D(p,q)< \tau_h$. We illustrate this procedure in Figure~\ref{fig:hd_nms}.

\subsection{Tracklet generation}

As stated by \cite{bise}, long trajectories obtained via frame-by-frame association may include more failures (such as drift and occlusions) than short trajectories. Hence, it is better to first reliably associate cell detections in adjacent frames,  and then use some global data association algorithm to obtain the final long-term tracklets. 

\begin{figure}[]
    \centering
    \includegraphics[width = 0.3\textwidth]{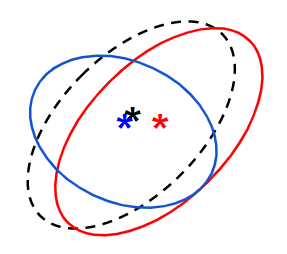}
    \caption[Illustration of using the Helinger Distance for associating detections in subsequent frames.]{Illustration of using the Hellinger Distance for associating detections in subsequent frames. The dashed ellipse (black) represents a detection in frame $i$, while the two solid ellipses (red and blue) represent two possible associations in frame $i\!+\!1$ (the stars mark the respective centers of the detections).     The Hellinger distance eliminates the imprecision of joining the black and blue detections due to only relating their centroid distances, while the red and black share more shape similarity.}
    \label{fig:hd_hungarian}
\end{figure}

In this work, ``reliable tracklets'' are obtained by computing the overlap between all detections of subsequent frames with the Hellinger distance (Eq.~\eqref{eq:helinger}), and the optimal association between detections is obtained by solving a linear sum assignment problem with the Hungarian Algorithm~\cite{kuhn1955hungarian}. Note that $H_D$ jointly considers the centroid distance (typically used for tracklet generation) and shape information -- encoded in the covariance matrix. Hence, nearby cells with distinct shape/orientations that generate strong ambiguity when using only the centroid distance can be disambiguated through the Hellinger distance (see Figure~\ref{fig:hd_hungarian}).

In order to avoid bad associations caused by false positive detections, we only associate pairs of cells $p$ and $q$ for which $H_D(p,q)<\tau_o$, where $\tau_o$ is a similarity threshold. Note that we implicitly assume a low cell displacement and shape change in adjacent frames, which is typically the case for time-lapse microscopy imagery~\cite{bise}. 

\subsection{Global data association}

In an ideal scenario, the combination of the associations in adjacent frames would lead to the long-term track of each cell. However, there may be errors either in cell detection or the association process itself. Furthermore, we also have to consider cell division and death. Hence, we explore a global data association step to fill the gaps between disjointed tracklets, remove false positives and identify the mitoses events. 

We based our method on the work by \cite{bise}, who formulated a maximum-a-posteriori problem (MAP) solved by linear programming that addresses the tree structure association problem. We proceed to briefly describe their method and then introduce our modifications. The MAP problem is solved by defining a set of hypotheses associated with a likelihood score for combinations of the $N_T$ tracklets (i.e., each tracklet will respond to a set of possible hypotheses with their likelihood). More precisely, they assume the following five possible types of hypotheses that can be made for each tracklet: initiation, termination, translation, mitoses, and false positive. 

Mathematically, let  $\bm{C}_{M\times 2N_T}$ be a binary matrix containing the constraints for all possible hypotheses. Each row of $\bm{C}$ relates to a single hypothesis, and it presents $2N_T$ columns that indicate the possible tracklet associations, where the first $N_T$ columns indicate the index of the source tracklet index and the following $N_T$ columns relate to the target tracklet index (or indices). For example, the translation hypothesis presents one source tracklet and one target tracklet. Meanwhile, the mitosis hypothesis relates one source tracklet to a pair of target tracklets (its children), and the false positive hypothesis relates one tracklet to itself (i.e., the source index is the same as the target). There is also a likelihood value for each hypothesis (i.e., for each row of $\bm{C}$), stored in a vector $\bm{\rho}$ with $M$ elements (the number of generated hypotheses).

The solution of the global optimization problem is a subset of all \textit{non-conflicting hypotheses} (a subset of rows of $\bm{C}$) that maximizes the sum of the corresponding likelihoods. This can be formulated as the following ILP optimization problem:
\begin{equation}
    \bm{x}^* = \mathop{\text{argmax}}_{\bm{x}} \bm{\rho} ^T \bm{x},~~s.t.~,\bm{C}^T \bm{x} \leq \bm{1},
    \label{eq:MAP}
\end{equation}
where ${\bm{x}_{M\times 1}}$ is a binary vector, and an entry $x_k=1$ means the ${k\text{-th}}$ hypothesis is selected in the global optimal solution. The constraint ${\bm{C}^T \bm{x} \leq \bm{1}}$ guarantees that each tracklet ID appears in only one associated tree or false positive tracklet.

The work developed by \cite{bise} allows a simple but efficient method for long-term data association. Yet, their approach still presents some limitations that we propose to address with the following modifications:
\begin{enumerate}
    \item In \cite{bise}, they propose to use a specific algorithm~\cite{huh2010automated} to identify the mitotic cells (i.e., the cells that are more likely to suffer mitoses). This algorithm, however, is specifically designed to work with images captured under phase-contrast microscopy, which narrows its usage to other types of images (e.g., fluorescence). We decided not to differentiate between mitotic and non-mitotic cells, and consider all cells as potentially mitotic. Furthermore, to address this choice, we employ two free parameters to adjust the likelihood distribution of the translation and mitoses hypothesis;
    \item We removed the initiation and termination hypotheses and replaced them with a \emph{completeness} one. In theory, these two hypotheses are essential to define the tree structure of the MAP problem. However, we noted in our experiments that they tend to generate a more complex MAP problem and do not provide better tree structures (i.e., closer to the ground-truth one), as we demonstrate in Section~\ref{sec:results}. In the original formulation, these hypotheses are defined regarding the cell position towards the borders and their first (or last) appearance, assuming that cells appear or disappear when they enter or leave the area captured by the microscope. Although these assumptions work well for an ideal detector, real detectors can fail to capture some cells (e.g., due to the absence of contrast between cell and background). As a consequence,  the tracklets can start to be recognized only after some frames and/or with the cells already located farther away from the image boundaries, and imposing these boundary-related conditions can increase the number of false negatives;
    \item We re-defined all the likelihood computations to use only information regarding the detected cells (e.g., position, time-frame distance, and confidence score returned by the object detector). In particular, we explore the confidence score to discriminate between the true and false positive hypotheses. For the translation and mitoses hypotheses, we used only the center and time distance between detections, whereas generic feature matching (e.g., based on histogram matching) was proposed by \cite{bise}. Our motivation is that finding an adequate feature to individually discriminate cells in long-term matching is challenging (no specific feature was mentioned in \cite{bise}), and appearance-based features might change depending on the method used to capture the images. Moreover, using only positional information and confidence score allow using any cell detector. We also removed the \textit{true positive likelihood} that was originally proposed by \cite{bise} since our pre-processing step can remove low-scored detections.
\end{enumerate}

The proposed alternate hypotheses and corresponding likelihoods are explained next.
\begin{itemize}
    \item \textbf{Translation hypothesis:}\\
    If the time and center distances between the last detection of tracklet $X_{k1}$ and the first detection of $X_{k2}$ are smaller than a pair of thresholds, $X_{k1}\rightarrow X_{k2}$ is a candidate of a tracklet translation. Considering that $h$ denotes the index of a new hypothesis, we append a new row to $\bm{C}$ and a corresponding likelihood to $\bm{\rho}$ as:
    \begin{align*}
        &C(h,i) = 
        \begin{cases}
          1,&\text{if $i=k_1$ or $i=N_T+k_2$}\\
          0,&\text{otherwise}
        \end{cases}, \\
        &\rho (h) = P_{link}(X_{k2}|X_{k1}).
    \end{align*}
    
    \item \textbf{Mitosis hypothesis:}\\
    If the time and center distances between the last detection of tracklets $X_{p}$ and the first detection of $X_{c1}$ and $X_{c2}$ are smaller than a threshold for the detection center and time, $X_{p}\rightarrow \{X_{c1},X_{c2}\}$ is a candidate of a tracklet mitosis. We define new entries for $\bm{C}$ and $\bm{\rho}$ as:
    \begin{align*}
        &C(h,i) = 
        \begin{cases}
          1,&\text{if $i=p$ or $i=N_T+c_1$ or $i=N_T+c_2$}\\
          0,&\text{otherwise}
        \end{cases}, \\
        &\rho (h) = P_{mit}(X_{c1},X_{c2}|X_p).
    \end{align*}
    
    \item \textbf{False positive and Completeness hypothesis:}\\
    If the score of tracklet $X_k$ (i.e., the mean score of all its detections) is smaller than threshold $\tau_{FP}$, $X_k$ is a candidate for both false positive and completeness. We define two new entries for $\bm{C}$ ($h$ and $h+1$) and $\bm{\rho}$ given by
    \begin{align*}
        &C(h,i) = C(h+1,i) = 
        \begin{cases}
          1,&\text{if $i=k$ or $i=N_T+k$}\\
          0,&\text{otherwise}
        \end{cases},
    \end{align*}
    and for the $\bm{\rho}$ entries as:
    \begin{align*}
        &\rho (h) = P_{FP}(X_k),\\
        &\rho (h+1) = P_{cplt}(X_k).
    \end{align*}
    
    The completeness hypothesis aims to cover the cases in which a tracklet does not translate or suffer from mitoses, i.e., the tracklet is, in fact, a full long-term cell track. Hence, assuming only the false positive hypothesis would wrongly eliminate those tracklets.
\end{itemize}

Now, we formalize the likelihoods for each hypothesis provided above. For the link and mitosis hypotheses, we would like to maximize the likelihood (i.e., values closer to 1) when the time and space distance between two cell tracklets are small, and minimize it (i.e., values closer to 0) when their distance increase. For this purpose, we based the formulation of the link and mitoses likelihoods on an exponential mapping that penalizes the time-space distance between tracklets. More precisely, we propose to use
\begin{equation}
    P_{link}(X_j|X_i) = \text{exp}\left(-\frac{(c_{i,j}+1)~t_{j,i}}{\Delta t~\lambda_{link}} \right),
\end{equation}
where $\Delta t$ is the dataset capture time step in $\frac{\text{frames}}{\text{hour}}$, $c_{i,j}$ is the Euclidean center distance between the last and the first detection of tracklets $X_i$ and $X_j$, respectively, $t_{i,j}$ is their time distance in frames, and $\lambda_{link}$ is a parameter that controls the decay of the exponential.

The mitosis likelihood is defined in a similar way, but jointly considering the space-time distance between the parent cell $p$ and the two candidate children $c1$, $c2$. Formally, it is given by
\begin{equation}
    \small
    P_{mit}(X_{c1},X_{c2}|X_p) = \text{exp}\left(-\frac{(c_{p,c1}+c_{p,c2}+1)(t_{p,c1}+t_{p,c2})}{4\Delta t~\lambda_{mit}} \right),
    \label{abc}
\end{equation}
where $\lambda_{mit}$ controls the decay of the exponential. We do not use the Helinger distance in these formulations because cell shapes can change considerably in longer-term translations or during mitosis.

For the false positive likelihood, we also explore the precision $\alpha$ of the object detector computed for a given dataset (e.g., the training or validation set) and the tracklet score $s_i$, defined as the mean confidence score of all its detection. We want low-confidence tracklets (w.r.t. to the detector precision) to have an increased false positive likelihood, but longer tracklets must have a smaller value since they tend to be related to actual tracklets. Based on these assumptions, the false positive likelihood is computed as
\begin{equation}
    P_{FP}(X_i) = (1-\alpha)(1-s_i+\tau_s)^{|X_i|},
\end{equation}
where $|X_i|$ is the number of total detection responses in the tracklet, and $\tau_s$ is the threshold detection score defined in Section~\ref{sec:detection_filtering}. Note that detectors with low precision (${\alpha<\!\!<1}$) are prone to produce false positives, and $P_{FP}$ should be increased in this case. On the other hand, detectors with high precision (${\alpha \approx 1}$) are less prone to produce false positives, and the likelihood $P_{FP}$ is decreased.

A similar rationale is used to define the completeness tracklet likelihood, given by
\begin{equation}
    P_{cplt}(X_i) = \alpha (s_i - \tau_s).
\end{equation}
Note that the likelihood of a tracklet being complete (i.e., it does not fit any other hypotheses) is not directly related to the tracklet size, since a cell can emerge and die very soon or appear in the visible field only in the last frames, which might lead to very small tracklets. Hence, we chose to define the likelihood based only on how precise the predictions of the object detector are, so that the tracklet generation step can do most of the correct associations between adjacent frames.

\begin{figure}[]
    \centering
    \includegraphics[width=.9\textwidth]{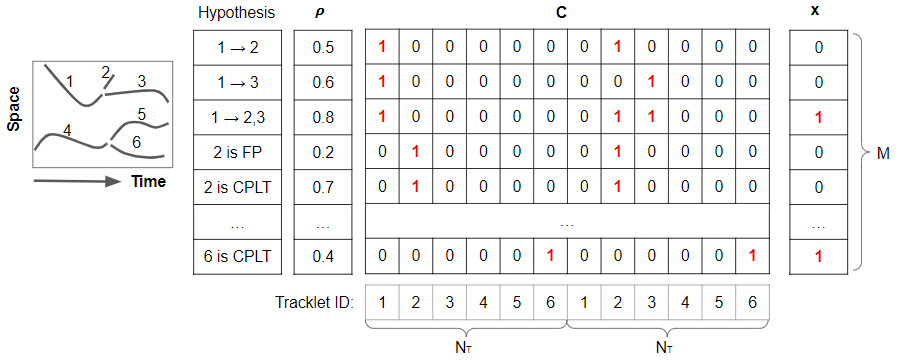}
    \caption[Example of hypotheses generation for a given set of tracklets.]{Example of hypotheses generation for a given set of tracklets. Given a set of initial tracklets, we generate hypothesis for each of them, which are associated with a likelihood value stored in vector $\bm{\rho}$, and the constraints stored in the binary matrix $\bm{C}$. Solving the MAP problem returns the binary vector $\bm{x}$ which selects a subset of rows in $\bm{C}$ that defines the optimal solution of the MAP problem. For example, the first line defines a translation hypothesis of tracklet 1 to 2 with a likelihood of 0.5. The matrix $\bm{C}$ stores the ID values of the considered tracklets (in each of its halves), and the value 0 in the vector $\bm{x}$ indicates that this hypothesis was not chosen during the optimization solving.}
    \label{fig:ILP_example}
\end{figure}

Figure~\ref{fig:ILP_example} shows an example of the proposed hypotheses generation. From left to right, it shows a set of tracklets; the possible hypotheses connecting the tracklets; the likelihood vector $\bm{\rho}$; the matrix $\bm{C}$ indicating the possible tracklet connections; and the vector $\bm{x}$ returned by the MAP problem solution. The final tracks are then obtained by solving Eq.~\eqref{eq:MAP}.

\section{Experiments}

\subsection{Datasets}\label{sec:datasets}

\begin{figure*}[]
\centering
\subfloat[Fluo-N2DH-\textbf{GOWT1}]{\includegraphics[width = 0.3\textwidth]{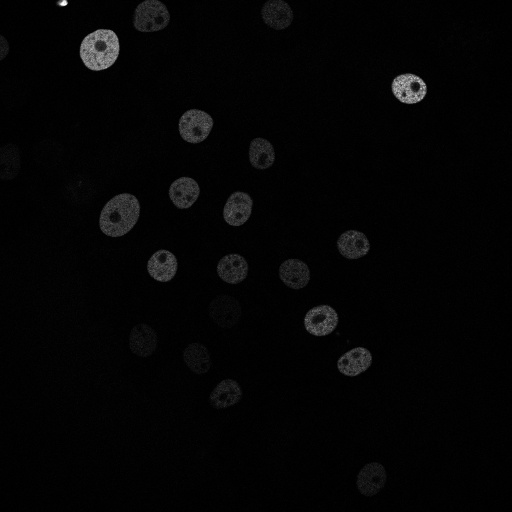}\label{fig:dataset:GOWT1}}~
\subfloat[Fluo-N2DH-\textbf{HeLa}]{\includegraphics[width = 0.3\textwidth]{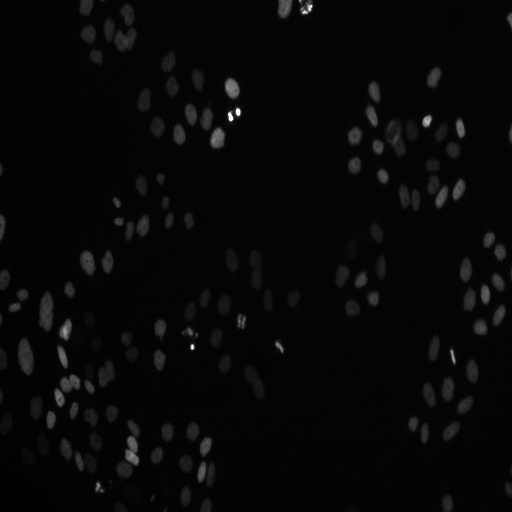}\label{fig:dataset:HELA}}~
\subfloat[PhC-C2DH-\textbf{U373}]{\includegraphics[width = 0.3\textwidth]{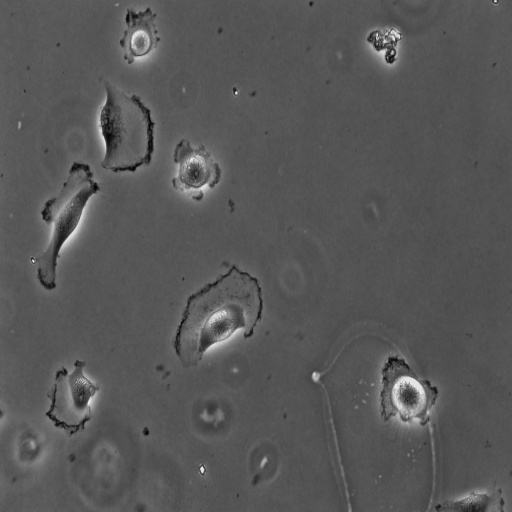}\label{fig:dataset:U373}}
\caption{Example of cell images for each of the datasets used for evaluation.}
\label{fig:example_datasets}
\end{figure*}

We evaluated the proposed method on three publicly available cell microscopy datasets provided from the ISBI 2015 Cell Tracking Challenge~\cite{isbi}: Fluo-N2DH-GOWT1, PhC-C2DH-U373, and Fluo-N2DL-HeLa, illustrated in Fig.~\ref{fig:example_datasets}. Each dataset contains two sequences in the training set (with ground truth annotations) and two challenge sequences (without ground truth annotation), named with suffices -01 and -02. The results for the challenge sequences are obtained by submitting the results to the ISBI challenge server. We proceed to provide details regarding each of the chosen datasets.

Fluo-N2DH-\textbf{GOWT1} contains GFP-transfected GOWT1 mouse embryonic stem cells captured on fluorescence microscopy. Challenges with this dataset include low contrast of some cells and few cells entering and exiting the imaged region from the axial direction. The capture time step is $\Delta t = 12 \frac{\text{frames}}{\text{hour}}$.

PhC-C2DH-\textbf{U373} contains glioblastoma-astrocytoma U373 cells captured under phase contrast microscopy. This dataset is challenging due to cells having highly deformable shapes and parts of cell bodies having a similar appearance to the background. The capture time step is $\Delta t = 4 \frac{\text{frames}}{\text{hour}}$.

Fluo-N2DH-\textbf{HeLa} contains fluorescently labeled HeLa nuclei captured on fluorescence microscopy. Challenges with this dataset include high cell density, low contrast, a few irregularly shaped cells, various mitoses events, and cells entering and exiting the imaged region. The capture time step is $\Delta t = 2 \frac{\text{frames}}{\text{hour}}$.

\begin{figure}[]
\centering
\subfloat[GOWT1-01]{\includegraphics[width = 0.25\textwidth]{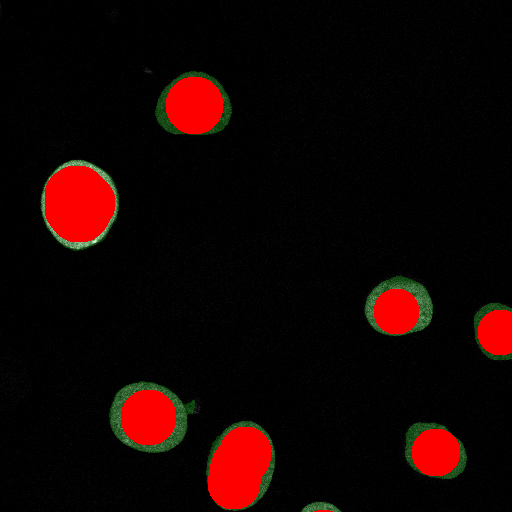}} ~
\subfloat[HeLa-01]{\includegraphics[width = 0.25\textwidth]{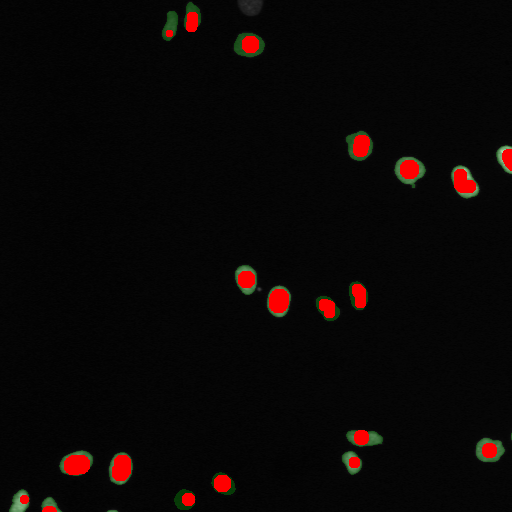}} ~
\subfloat[U373-01]{\includegraphics[width = 0.25\textwidth]{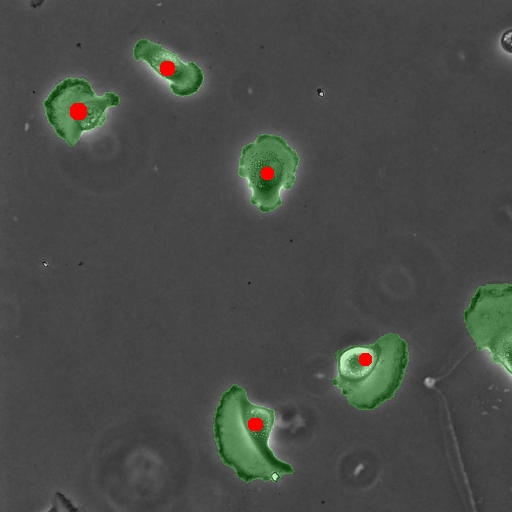}} \vspace{0.1cm} \\
\subfloat[GOWT1-02]{\includegraphics[width = 0.25\textwidth]{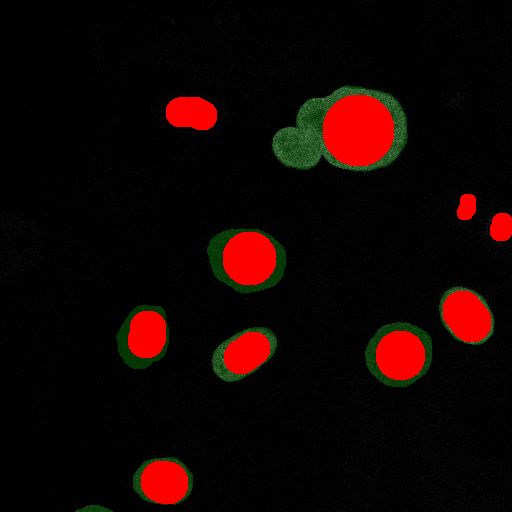}} ~
\subfloat[HeLa-02]{\includegraphics[width = 0.25\textwidth]{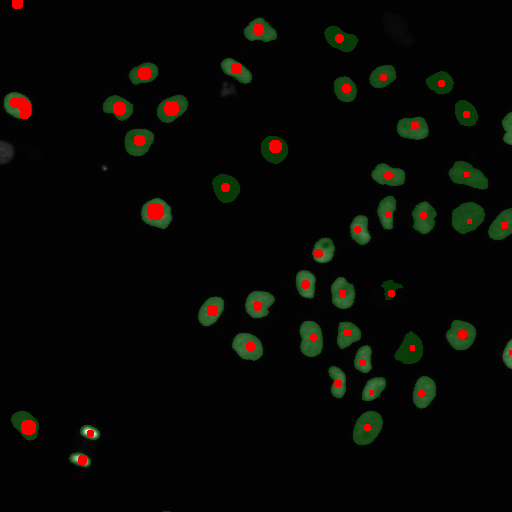}} ~
\subfloat[U373-02]{\includegraphics[width = 0.25\textwidth]{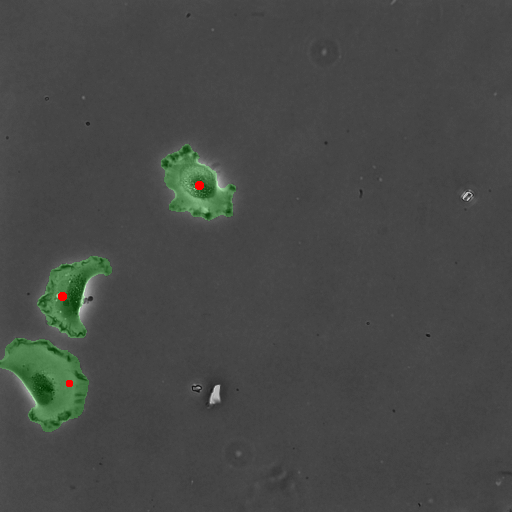}}
\caption[Illustration showing the difference between the two CTC annotations types.]{Illustration showing the difference between the two CTC~\cite{isbi} annotations types: cell masks (green) and cell markers (red).}
\label{fig:comparsion_masks_marks}
\end{figure}

All these datasets only contain ground truth (GT) annotations for cells within a field of interest, which excludes a few pixels for cells close to the image boundaries. There are two types of GT annotations: cell masks for the segmentation evaluation, and cell markers for the detection and tracking evaluation. For the cell masks, the annotations are provided as \emph{silver} and \emph{gold} standards. The silver standard annotations refer to computer-origin reference annotations, while the gold standard refers to human-origin ones. Since only a few cells are annotated in the gold standard, we used only the silver ones for both training and evaluation. The cell marks are ``similar'' to the segmentation masks, but they have reduced size and serve solely as a position descriptor of the cells. Moreover, they do not follow a standard regarding their size or placement on the cell image. We illustrate these annotation discrepancies in Fig.~\ref{fig:comparsion_masks_marks}.

\subsection{Data pre-processing} \label{sec:data_handling}

We use the same data pre-processing procedure for all datasets, except for some specific adjustments in the HeLa, as described next. Since the HeLa dataset contains cells with very different sizes compared to the other CTC datasets~\cite{isbi}, we followed a similar strategy to the one employed by \cite{cpn} and magnified all images by a factor of 2. This procedure allows us to use the same network architecture without needing any adjustment in its head parameters regressor (e.g., the anchors) to better fit the cell sizes in this dataset.

For training the object detector, we extracted patches of  full images in the datasets using a $512\!\times\!512$px sliding window with $100$px stride. The patches are extracted starting the window at the top-left corner of the image and sliding it across the image horizontally and vertically according to the specified stride. At each position, we extract a patch within the window boundaries. This process resulted in, for each sequence of the CTC dataset, $4,\!508$ for the GOWT1 dataset, $690$ for U373, and $16,\!314$ for HeLa. During inference, we used the full-sized images, except for the HeLa dataset, for which we used the patches with a $256$ stride (to provide some overlap in the borders) and then divided the parameters of the detection by a factor of 2 to retrieve them with the expected original image sizes. We used a similar strategy as in other works \cite{kth,epflheid,blob,cpn,drl}: with dataset sequences that contain GT annotations (i.e., not the one related to the challenge), we trained the models with one sequence and used the other for evaluation.   
To generate the OBBs (and then the EBBs) from the CTC cell masks, which are required to train the object detectors, we employed the minimum-area rectangle fitting algorithm available in the OpenCV~\cite{opencv} framework.

\subsection{Implementation details}

As the object detector, we chose to use the R2CNN~\cite{r2cnn} provided by the AlphaRotate benchmark~\footnote{\url{https://github.com/yangxue0827/RotationDetection}}, because it has shown to have the best compromise between precision and recall compared to other OBB detectors in our preliminary investigations using a private cell dataset. Furthermore, as presented in \cite{liu2020deep}, general-purpose two-stage detectors usually provide better results when compared with one-stage ones. R2CNN is a two-stage object detector (i.e. it has a region proposed module (RPN) before the detection module) that was first proposed in text detection problems. The model parameters used for all experiments are the same as in the AlphaRotate default architecture. The trained model uses a ResNet50~\cite{resnet} backbone with pre-trained weights on the ImageNet~dataset~\cite{imagenet}. Both classification modules (RPN and head) use the categorical cross-entropy loss, and both box regression modules use the smooth-$\ell_1$ loss. Weight decay and momentum are set to $10^{-4}$ and 0.9, respectively. We employ Momentum Optimizer over 1 GPU and eight images per mini-batch. All models are trained for 100 epochs with a 0.1 reduction factor of the learning rate at epochs 12, 16, and 20, using random rotation and flips as data augmentation primitives. The initial learning rate was $10^{-3}$ for all datasets except HeLa, for which the model showed overfitting. In this dataset, we used $10^{-4}$ as the initial learning rate and trained for 24 epochs only.

The hyper-parameters for the tracking system are the same for all datasets (except the object detection precision $\alpha$  and capture time $\Delta t$), and were chosen empirically to produce good results for all the evaluated datasets (i.e., we used the training datasets in order to define the hyper-parameters of the final tracking system that was evaluated on the CTC~\cite{isbi} server with hidden GT). Nevertheless, we provide a sensitivity analysis of the parameters in Section~\ref{sec:abblation}, and conclude that changing them has a small impact on the results. For the inference of the detector, we used a score threshold $\tau_s=0.5$, and an overlap threshold of $\tau_h=0.5$ for the filtering and aggregation step. For the tracklet generation, we employed an overlap threshold of $\tau_o=0.5$, which are classical thresholds for IoU-like metrics.

For the parameters of the global data association algorithm, we used a time threshold $t_{th}=3$ frames, and a false positive threshold $\tau_{FP}=0.9$. The space threshold was set to $0.1\sqrt{W_f^2+H_f^2}$, where $W_f$ and $H_f$ are the width and height of the dataset frames, respectively. The $\alpha$ value was computed using the precision value for each dataset considering an overlap threshold of $\tau_h=0.5$ between the predicted and ground-truth detections of the training images with no score threshold (i.e., considering any detection with a confidence score above zero), and are available in Table~\ref{tab:alpha_values}. The free parameters for likelihood adjustment are set to $\lambda_{link}=25$ and $\lambda_{mit}=50$ for all datasets. The MAP problem was solved using the Cbc~\cite{cbc} mixed ILP solver provided in the CVXPY\footnote{Available at: \url{https://www.cvxpy.org/index.html}} Python~3~\cite{python} library.

\begin{table}[]
    \caption{Object detection precision $\alpha$ for each dataset.}
    \centering
    \begin{tabular}{c|c|c|c|c|c|c}
        \textbf{Dataset} & GOWT1-01 & GOWT1-02 & U373-01 & U373-02 & HeLa-01 & HeLa-02 \\ \hline
        $\bm{\alpha}$ &  0.8993 & 0.8644 & 0.7673 & 0.6867 & 0.7930 & 0.7899
    \end{tabular}
\label{tab:alpha_values}
\end{table}

\subsection{Evaluation Metrics}\label{sec:metrics}

We evaluated our method using the metrics proposed by the ISBI 2015 Cell Tracking Challenge~\cite{isbi} and standard literature metrics used for evaluating general-purpose detection and tracking systems. The ISBI Challenge provides the DET, SEG and TRA metrics~\footnote{A full description of the metrics can be found at \url{http://celltrackingchallenge.net/evaluation-methodology/}}. Both DET and TRA metrics are designed to mirror the manual effort required to correct the errors of a given detection and tracking algorithm, respectively, using Acyclic Oriented Graph Matching; SEG measures the Jaccard similarity index (a.k.a. IoU) between predicted and ground-truth segmentation masks. All ISBI metrics return values from 0 to 1 (1 being the highest score). 

Since our method provides only an approximation of the segmentation masks through EBBs, we would also like to estimate how close both the GT EBBs and the predicted EBBs are to the GT segmentation masks. The former can be answered by evaluating the EBBs generated from the ground truth cell masks with the SEG metric, obtaining the \emph{EBB SEG} metric. The latter is obtained by simply evaluating our method with the SEG metric.

\begin{figure}[]
\centering
\includegraphics[width = 0.35\textwidth]{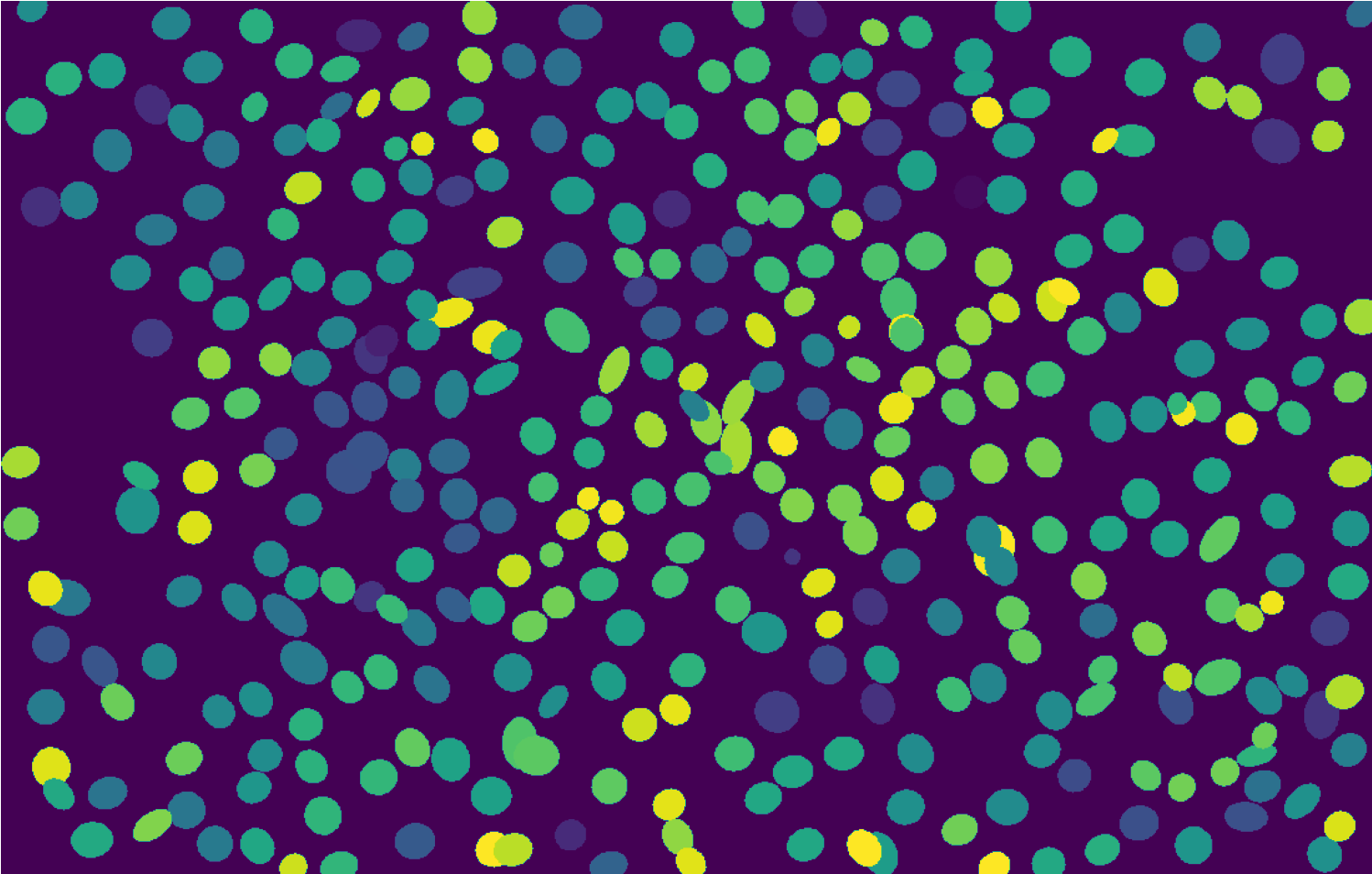} ~
\includegraphics[width = 0.35\textwidth]{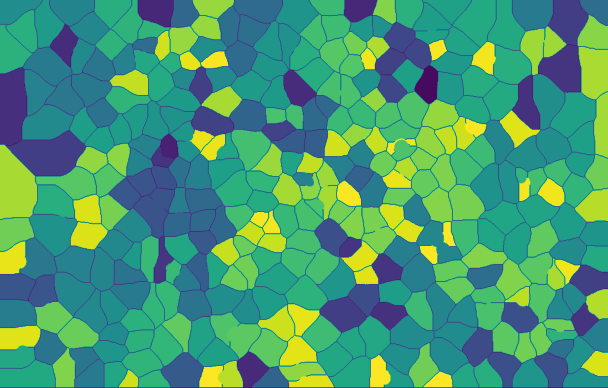}
\caption{Visual comparison of the cell masks using EBBs (left) and post-processed with watershed (right). Note that the watershed fills the voids between the different detections allowing some degree of error when evaluating with the DET and TRA metrics.}
\label{fig:watershed}
\end{figure}

For the detection and tracking evaluation, the algorithm provided by the ISBI challenge disregards detections that do not entirely overlap with the provided ground-truth cell marks. As mentioned before, they do not follow a standard for size and displacement, which might affect the quantitative metrics. For the  U373-02 dataset, in particular,  Ulah and colleagues \cite{cpn} propose simply enlarging the predicted masks to avoid missing the cell marks. In our approach, however, the variability of cell mark annotations can significantly impact all the tested datasets, since the EBBs are only approximations of the segmentation mask and might not completely overlap with a ground-truth mark -- they are not bounding representations as well. In order to overcome this issue, we also evaluated our method enlarging the predicted cell masks with a simple watershed algorithm, using the EBBs as guiding markers. An example of EBB-guided watershed is shown in Fig.~\ref{fig:watershed}. 

For the standard literature metrics, we evaluated our object detector using the Average Precision (AP) metric, which relates the precision and recall for a given threshold score and detection overlap (e.g., using IoU or the Hellinger distance) between detections. For the tracking evaluation, we used the CLEAR-MOT~\cite{mota,mot16} and ID-MEASURE~\cite{mota_id} metrics\footnote{We used a Python~\cite{python} implementation of these metrics available at: \url{https://github.com/cheind/py-motmetrics}}. Both metrics attempt to find a minimum-cost assignment between ground truth objects and predictions. However, while CLEAR-MOT solves the assignment problem on a local per-frame basis, ID-MEASURE solves the bipartite graph matching by finding the minimum cost of objects and predictions over all frames. For evaluating the object detector, we used the Hellinger distance to compute the overlap between predicted and ground-truth detections, and set the score and overlap thresholds to 0.5.

\section{Results and Discussions}\label{sec:results}

In this section, we present the results for evaluating our tracking-by-detection method in the CTC~\cite{isbi} datasets. We evaluated our results in two manners: one by using only the training datasets with GT and comparing with other works directly (see Section~\ref{sec:data_handling} for more details), and the other by submitting our codes to the CTC server\footnote{Available at: \url{http://celltrackingchallenge.net/}} in order to retrieve the metrics scores and rank compared to other submitted works. We begin this section by briefly describing the baseline methods used in the first evaluation methodology. Then, we present the results for both methodologies and discuss them.

\subsection{Baseline methods}

We proceed to briefly describe the methods used as baseline for comparison with our proposal, highlighting the nature of required training data.

\paragraph{KTH}~\cite{kth} segments cells using a bandpass filter followed by thresholding, and then uses the watershed algorithm to split joined cells. The tracking graph is created by connecting cell segmentations in adjacent frames, and then solved by iteratively finding the lowest cost path in the graph using Viterbi algorithm. It does not require any annotation.

\paragraph{EPFL}~\cite{epflheid} detects cells by fitting ellipses to binary segmented regions. It joins the detection on subsequent frames using a tracking graph solved using integer linear programming (ILP). It requires annotation for segmentation, detection and tracking.

\paragraph{HEID}~\cite{epflheid} detects cells by merging super-pixels clustering segmentation, which are obtained using watersheds. Then, the tracking is retrieved by finding the global optimum of a graph model that represents cellular events using ILP. It requires annotation for segmentation, detection, and tracking.

\paragraph{BLOB}~\cite{blob} detect cells using multiple elliptical filter banks, and performs tracking by iteratively finding the shortest path in a model graph. It requires only tracking annotation, i.e., temporal cell associations.

\paragraph{CPN}~\cite{cpn} first generates cell region proposals using an HBB object detector, and then finds the segmentation masks of these regions using a deep learning segmentation network similar to the U-Net~\cite{unet}. It performs tracking using ILP to solve a graph for which the weights are set by training a random forest classifier with several histogram features. It requires full annotation for detection, segmentation, and tracking.

\paragraph{DRL}~\cite{drl} uses the U-Net~\cite{unet} model to produce the cell segmentation masks. Then, it uses deep reinforcement learning to build the cost matrix that joins the cells of subsequent frames. It requires full annotation for detection, segmentation, and tracking.

\paragraph{ST-TCV}~\cite{boukari2018automated} detects cells using a joint spatio-temporal diffusion and region-based level-set optimization approach~\cite{boukari2016joint}. Then, it uses motion prediction and minimization of a global probabilistic function to join the cells of subsequent frames. It does not require any annotation.

We did not include results for the methods U-Net~\cite{unet}, GC-ME~\cite{gcme}, and U-Net~S~\cite{unets} because they do not follow the same data split methodology (i.e., they employ images from both sequences in the training phase), and hence it would define an unfair baseline for comparison.

\subsection{Results for the CTC datasets}
The results for comparing our method with SOTA approaches on the ISBI evaluation using separate sets are provided in Table~\ref{tab:ISBI_results}. We include in this table results for both using or not using the watershed algorithm for enlarging the detection results.

\begin{table}[]
\centering
\caption[Results for the CTC training datasets using separate sequences.]{Results for the CTC~\cite{isbi} training datasets using separate sequences. Our results with the watershed method are reported as \emph{Ours-W}. \textbf{Bold} values mark the best results, while \underline{underline} values mark the second best. We also report the annotation requirements (Ann.~Req. column) of each technique related to the detection (Det) and tracking (Tra)}.

\begin{tabular}{l|l|ll|llll}
\multicolumn{1}{c|}{} & \multicolumn{1}{c|}{} & \multicolumn{2}{c|}{Ann. Req.} & \multicolumn{1}{c}{} & \multicolumn{1}{c}{} & \multicolumn{1}{c}{} & \multicolumn{1}{c}{} \\ \cline{3-4}
\multicolumn{1}{c|}{\multirow{-2}{*}{Dataset}} & \multicolumn{1}{c|}{\multirow{-2}{*}{Method}} & \multicolumn{1}{c}{Det.} & \multicolumn{1}{c|}{Tra.} & \multicolumn{1}{c}{\multirow{-2}{*}{DET}} & \multicolumn{1}{c}{\multirow{-2}{*}{SEG}} & \multicolumn{1}{c}{\multirow{-2}{*}{EBB SEG}} & \multicolumn{1}{c}{\multirow{-2}{*}{TRA}} \\ \hline

 & ST-TCV & \xmark & \xmark & \textit{N/A} & \textit{N/A} & - & 0.913 \\
 & KTH & \xmark & \xmark & \textit{N/A} & 0.6849 & - & 0.9462 \\
 & BLOB & \xmark & \cmark & \textit{N/A} & 0.7415 & - & 0.9733 \\
 & CPN & \cmark & \cmark & \textit{N/A} & 0.8506 & - & 0.9864 \\
 & DRL & \cmark & \cmark & \textit{N/A} & \textbf{0.8585} & - & 0.9875 \\
 & \textbf{Ours} & \cmark & \xmark & 0.9916 & \underline{0.8568} & 0.9268 & \underline{0.9914} \\
\multirow{-9}{*}{GOWT1-01} & \textbf{Ours-W} & \cmark & \xmark & 0.9940 & - & - & \textbf{0.9930}\\ \hline

 & ST-TCV & \xmark & \xmark & \textit{N/A} & \textit{N/A} & - & 0.914 \\
 & HEID & \cmark & \cmark & \textit{N/A} & \textit{N/A} & - & 0.95 \\
 & EPFL & \cmark & \cmark & \textit{N/A} & \textit{N/A} & - & 0.95 \\
 & KTH & \xmark & \xmark & \textit{N/A} & 0.8942 & - & 0.9452 \\
 & BLOB & \xmark & \cmark & \textit{N/A} & \underline{0.9046} & - & 0.9628 \\
 & CPN & \cmark & \cmark & \textit{N/A} & 0.8725 & - & 0.9719 \\
 & DRL & \cmark & \cmark & \textit{N/A} & \textbf{0.9181} & - & 0.9575 \\
 & \textbf{Ours} & \cmark & \xmark & 0.9812 & 0.8509 & 0.9167 & \underline{0.9817} \\
\multirow{-9}{*}{GOWT1-02} & \textbf{Ours-W} & \cmark & \xmark & 0.9868 & - & - & \textbf{0.9853}\\ \hline

 & CPN & \cmark & \cmark & \textit{N/A} & \underline{0.7336} & - & 0.9594 \\
 & DRL & \cmark & \cmark & \textit{N/A} & \textbf{0.8527} & - & \textbf{0.9919} \\
 & \textbf{Ours} & \cmark & \xmark & 0.9647 & 0.6307 & 0.7791 & 0.9671 \\
\multirow{-4}{*}{U373-01} & \textbf{Ours-W} & \cmark & \xmark & 0.9748 & - & - & \underline{0.9774} \\ \hline

 & CPN & \cmark & \cmark & \textit{N/A} & \underline{0.7376} & - & \underline{0.9346}* \\
 & DRL & \cmark & \cmark & \textit{N/A} & \textbf{0.7735} & - & 0.9318 \\
 & \textbf{Ours} & \cmark & \xmark & 0.8822 & 0.5626 & 0.7029 & 0.8737 \\
\multirow{-4}{*}{U373-02} & \textbf{Ours-W} & \cmark & \xmark & 0.9634 & - & - & \textbf{0.9525} \\ \hline

 & ST-TCV & \xmark & \xmark & \textit{N/A} & \textit{N/A} & - & 0.816 \\
 & HEID & \cmark & \cmark & \textit{N/A} & \textit{N/A} & - & 0.80 \\
 & EPFL & \cmark & \cmark & \textit{N/A} & \textit{N/A} & - & 0.98 \\
 & KTH & \xmark & \xmark & \textit{N/A} & 0.8018 & - & 0.9775 \\
 & CPN & \cmark & \cmark & \textit{N/A} & \textbf{0.8313} & - & \textbf{0.9869} \\
 & BLOB & \xmark & \cmark & \textit{N/A} & \underline{0.7951} & - & 0.9803 \\
 & \textbf{Ours} & \cmark & \xmark & 0.9779 & 0.7264 & 0.8871 & 0.9758 \\
\multirow{-8}{*}{HeLa-01} & \textbf{Ours-W} & \cmark & \xmark & 0.9863 & - & - & \underline{0.9820} \\ \hline

 & ST-TCV & \xmark & \xmark & \textit{N/A} & \textit{N/A} & - & 0.845 \\
 & HEID & \cmark & \cmark & \textit{N/A} & \textit{N/A} & - & 0.85 \\
 & EPFL & \cmark & \cmark & \textit{N/A} & \textit{N/A} & - & 0.97 \\
 & KTH & \xmark & \xmark & \textit{N/A} & 0.8366 & - & 0.9747 \\
 & CPN & \cmark & \cmark & \textit{N/A} & \textbf{0.8445} & - & \textbf{0.9826} \\
 & BLOB & \xmark & \cmark & \textit{N/A} & \underline{0.8390} & - & \underline{0.9771} \\
 & \textbf{Ours} & \cmark & \xmark & 0.9707 & 0.7618 & 0.8897 & 0.9664 \\
\multirow{-8}{*}{HeLa-02} & \textbf{Ours-W} & \cmark & \xmark & 0.9796 & - & - & 0.9740
\end{tabular}

\begin{minipage}{\textwidth}
\vspace{0.1cm}
\small * denotes augmentation on the segmentation masks.
\end{minipage}
\label{tab:ISBI_results}
\end{table}

Regarding the approximation of the cell masks using EBBs, we can observe from the \emph{EBB SEG} column that it provides a good fit for the GOWT1 and HeLa datasets. On the other hand, the EBB approximation is not very good for the U373 dataset since there is strong variability in the cell shapes, as mentioned before. Finally, our method did not achieve SOTA scores on the SEG metric, which is expected since we only approximate the cell masks through EBBs. Nevertheless, it could reach close values to those on both the GOWT1 and HeLa datasets, and even get the second best for the GOWT1-01.

For the DET and TRA metrics, we note that our approach achieves a considerable boost using the watershed post-processing algorithm, particularly for the U373-02 dataset. We believe that this behavior is mostly due to the GT annotation of the cell markers in the dataset, which is sometimes located at the boundary of the cells and might not overlap completely with the EBB. Nevertheless, our method (without watershed) could reach SOTA results in both GOWT1 datasets, while having the second-best result in U373-01. When applying the watershed mask augmentation, our method reached SOTA scores on three datasets, and second best on other two. Regarding the degree of annotation required for each technique, our method was capable of outperforming fully supervised methods (HEID~\cite{epflheid}, EPFL~\cite{epflheid}, CPN~\cite{cpn} and DRL~\cite{drl}) in most of the evaluated datasets. It also presented better results than the tracking-supervised approach BLOB~\cite{blob} and the unsupervised trackers KTH~\cite{kth} and ST-TCV~\cite{boukari2018automated} in all datasets, except for HeLa-02 compared to the KTH method.

Table~\ref{tab:ISBI_challenge} report the results for evaluating our method on the CTC server~\footnote{Submitted implementation available at: \url{http://celltrackingchallenge.net/participants/UFRGS-BR/}} on the DET and TRA metrics with and without the watershed method\footnote{Due to environment problems related to CUDA instructions on the CTC server computers, we could not reproduce the exact same code used on our side. This ended up slightly harming the predicted EBB shapes and hence under-estimating the SEG metric, and the DET and TRA metrics when the detections' shapes are not augmented with the watershed algorithm.}. In this evaluation, our method was capable of achieving the TOP~3 rank on the DET metric for the GOWT1 dataset using the watershed algorithm. Although it could not overcome the SOTA in any dataset, we can observe a small difference between the scores of our method with those ranked as the top one. Furthermore, most of the top-ranked algorithms are end-to-end trackers or use elaborated techniques to improve the predicted segmentation masks from deep learning models (e.g., using model assemble or multiple refinement stages). In contrast, our proposed method intends to provide a simple yet efficient method for tracking-by-detection that requires only per-frame OBB cell annotations.

\begin{table}[]
\centering
\caption[Results from the CTC challenge evaluation server.]{Results from the CTC~\cite{isbi} challenge evaluation server. We report the results for our technique both using and not the watershed mask augmentation (\emph{W} column), the rank position over all submissions, and the relative difference to the first rank method (\emph{To TOP1} column). Evaluation date: October 10, 2022.}
\begin{tabular}{l|c|lll|lll}
\multirow{2}{*}{Dataset} & \multirow{2}{*}{W} & \multicolumn{3}{c|}{DET} & \multicolumn{3}{c}{TRA} \\ \cline{3-8} 
 &  & Score & Rank & To TOP1 & Score & Rank & To TOP1 \\ \hline
\multicolumn{1}{c|}{\multirow{2}{*}{GOWT1}} & \xmark & 0.925 & 26/49 & 5.61\% & 0.922 & 19/40 & 5.82\% \\
\multicolumn{1}{c|}{} & \cmark & 0.970 & 3/49 & 1.02\% & 0.959 & 4/40 & 2.04\% \\ \hline
\multirow{2}{*}{U373} & \xmark & 0.914 & 33/38 & 7.68\% & 0.909 & 26/31 & 7.72\% \\
 & \cmark & 0.979 & 17/38 & 1.11\% & 0.976 & 12/31 & 0.91\% \\ \hline
\multirow{2}{*}{HeLa} & \xmark & 0.986 & 15/48 & 0.80\% & 0.984 & 11/39 & 0.91\% \\
 & \cmark & 0.989 & 10/48 & 0.50\% & 0.988 & 8/39 & 0.50\%
\end{tabular}
\label{tab:ISBI_challenge}
\end{table}

The results using standard detection and quality metrics are provided in Table~\ref{tab:STD_results}. The detection results refer only to the detector performance itself, i.e., it does not use the global data association to further eliminate false positive detections and/or add false negative ones. This table evaluates different aspects of our method, enabling us to identify its strengths and weakness better. Regarding the recall metric, we can observe that it could achieve high values on all datasets for detection and tracking. However, it is noticeable that our method fails to eliminate false positive detections, which impact the precision in both detection and tracking, as noted for dataset U373-01. On the other hand, the detector precision in the GOWT1-02 dataset was also relatively low, but our global data association algorithm was capable of eliminating most of the false positive detections and hence obtaining a higher precision value on the tracking metrics. For U373-02, we can observe an inconsistency between the detection and tracking precision, which might be explained by the inconsistency of the cell mark annotations mentioned in Section~\ref{sec:metrics}.

\begin{table}[]
\caption[Results using standard detection and tracking quality metrics.]{Results using standard detection and tracking quality metrics (in \%), as described in Section~\ref{sec:metrics}.}
\centering
\begin{tabular}{l|llllll|llll}
\multicolumn{1}{c|}{\multirow{2}{*}{Dataset}} & \multicolumn{6}{c|}{Tracking} & \multicolumn{4}{c}{Detection} \\ \cline{2-11} 
\multicolumn{1}{c|}{} & ID-F1 & ID-P & ID-R & R & P & MOTA & R$_{50}$ & P$_{50}$ & F1$_{50}$ & AP$_{50}$ \\ \hline
GOWT1-01 & 97.9 & 96.7 & 99.0 & 99.9 & 97.5 & 97.0 & 99.4 & 97.0 & 98.2 & 90.9 \\
GOWT1-02 & 95.2 & 91.3 & 99.6 & 100.0 & 91.7 & 90.7 & 99.9 & 83.7 & 91.1 & 90.5 \\
U373-01 & 88.9 & 79.9 & 100.0 & 100.0 & 79.9 & 74.8 & 99.9 & 79.0 & 88.2 & 90.8 \\
U373-02 & 81.5 & 79.6 & 83.6 & 96.0 & 91.4 & 86.3 & 93.6 & 90.1 & 91.8 & 89.5 \\
HeLa-01 & 94.6 & 92.4 & 96.8 & 100.0 & 95.4 & 94.0 & 90.8 & 96.5 & 93.6 & 90.4 \\
HeLa-02 & 93.1 & 89.0 & 97.6 & 100.0 & 91.2 & 89.1 & 95.0 & 93.5 & 94.2 & 90.1
\end{tabular}
\label{tab:STD_results}
\end{table}

Table~\ref{tab:bise_comparison} shows a comparison of our tracking pipeline using our data association algorithm and a modified version using the approach by \cite{bise} for computing the final tracks. We note that the proposed modifications only slightly improve the DET and TRA metrics for most datasets, but they provide a significant reduction in the number of generated hypotheses. As a consequence, it allows faster solution computation and fewer hardware requirements.

\begin{table}[]
\centering
\caption[Results comparing our complete method and a modified version of \cite{bise}.]{Results comparing our complete method and a modified version of \cite{bise} (Baseline).}
\begin{tabular}{l|l|llll}
Dataset & Method & \multicolumn{1}{c}{DET} & \multicolumn{1}{c}{TRA} & Hypothesis & Time (s) \\ \hline
\multicolumn{1}{c|}{\multirow{2}{*}{GOWT1-01}} & Baseline & 0.9914 & 0.9913 & 133 & 0.1234 \\
\multicolumn{1}{c|}{} & Ours & 0.9916 & 0.9914 & 47 & 0.0998 \\ \hline
\multicolumn{1}{c|}{\multirow{2}{*}{GOWT1-02}} & Baseline & 0.9803 & 0.9805 & 312 & 0.2280 \\
\multicolumn{1}{c|}{} & Ours & 0.9812 & 0.9817 & 161 & 0.2082 \\ \hline
\multirow{2}{*}{U373-01} & Baseline & 0.9655 & 0.9677 & 78 & 0.0891 \\
 & Ours & 0.9647 & 0.9671 & 53 & 0.0763 \\ \hline
\multirow{2}{*}{U373-02} & Baseline & 0.8818 & 0.8733 & 83 & 0.0918 \\
 & Ours & 0.8822 & 0.8737 & 24 & 0.0509 \\ \hline
\multirow{2}{*}{HeLa-01} & Baseline & 0.9772 & 0.9741 & 7989 & 30.681 \\
 & Ours & 0.9779 & 0.9758 & 7061 & 25.337 \\ \hline
\multirow{2}{*}{HeLa-02} & Baseline & 0.9699 & 0.9652 & 25525 & 259.18 \\
 & Ours & 0.9707 & 0.9664 & 22649 & 235.06
\end{tabular}
\label{tab:bise_comparison}
\end{table}

\begin{figure}[]
\centering
\subfloat[GOWT1-01]{\includegraphics[width = 0.3\textwidth]{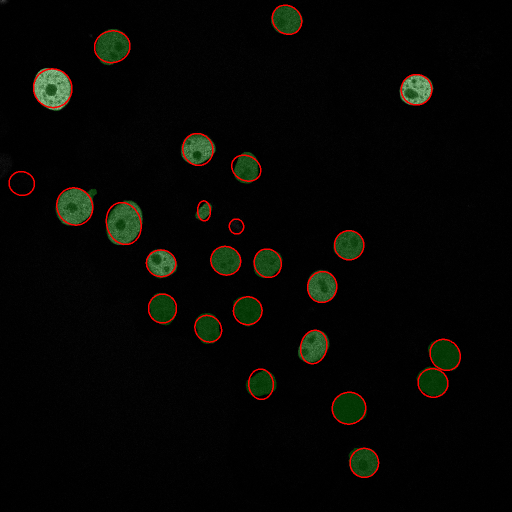}}~
\subfloat[GOWT1-02]{\includegraphics[width = 0.3\textwidth]{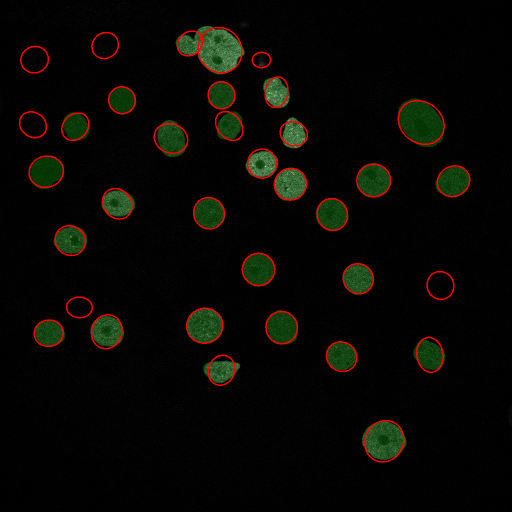}}\\
\subfloat[HeLa-01]{\includegraphics[width = 0.3\textwidth]{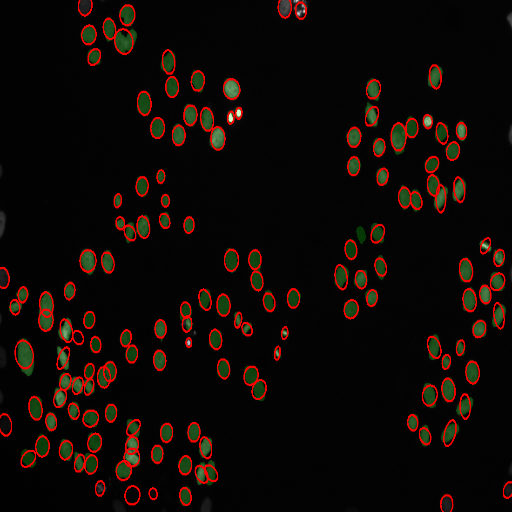}}~
\subfloat[HeLa-02]{\includegraphics[width = 0.3\textwidth]{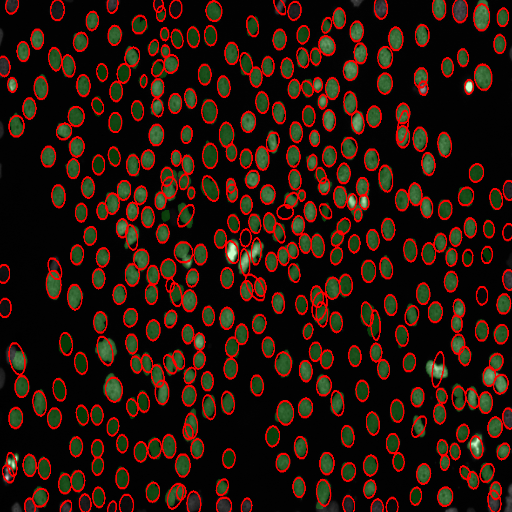}}\\
\subfloat[U373-01]{\includegraphics[width = 0.3\textwidth]{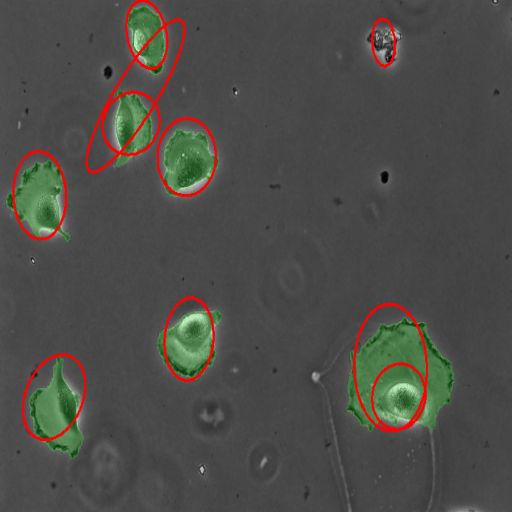}}~
\subfloat[U373-02]{\includegraphics[width = 0.3\textwidth]{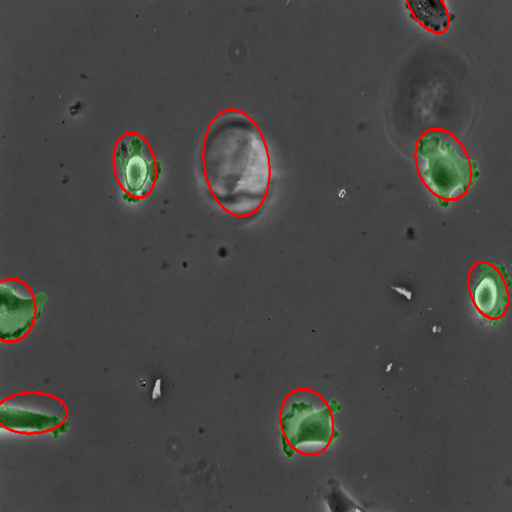}}\\
\caption[Visual results of our method in the CTC training datasets.]{Visual results of our method in the CTC~\cite{isbi} training datasets. In green are the ground-truth segmentation masks, and in red are the predicted EBBs.}
\label{fig:visual_results}
\end{figure}

\begin{figure}[]
\centering
\subfloat[GOWT1-01]{\includegraphics[width = 0.4\textwidth]{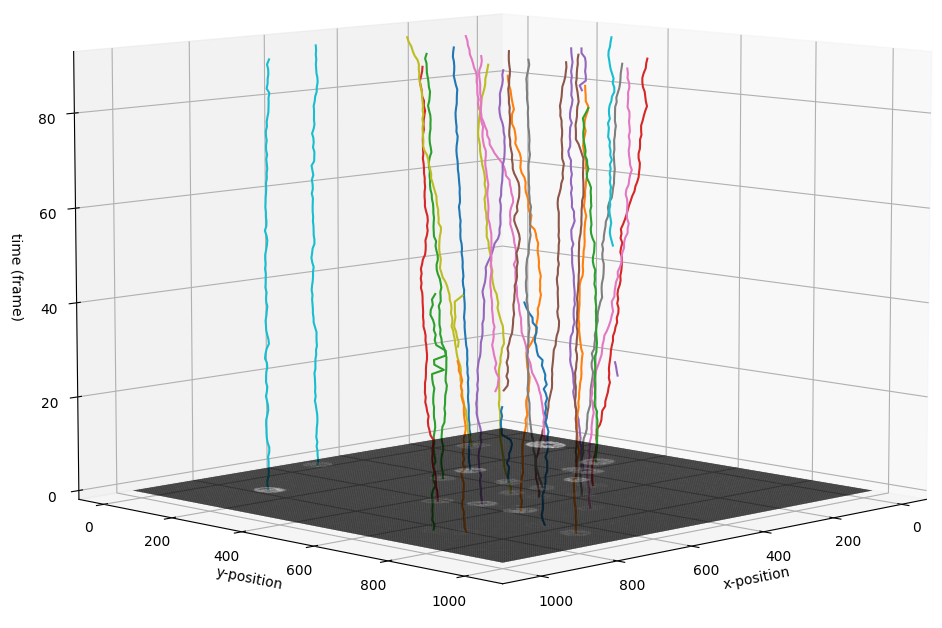}}~
\subfloat[GOWT1-02]{\includegraphics[width = 0.4\textwidth]{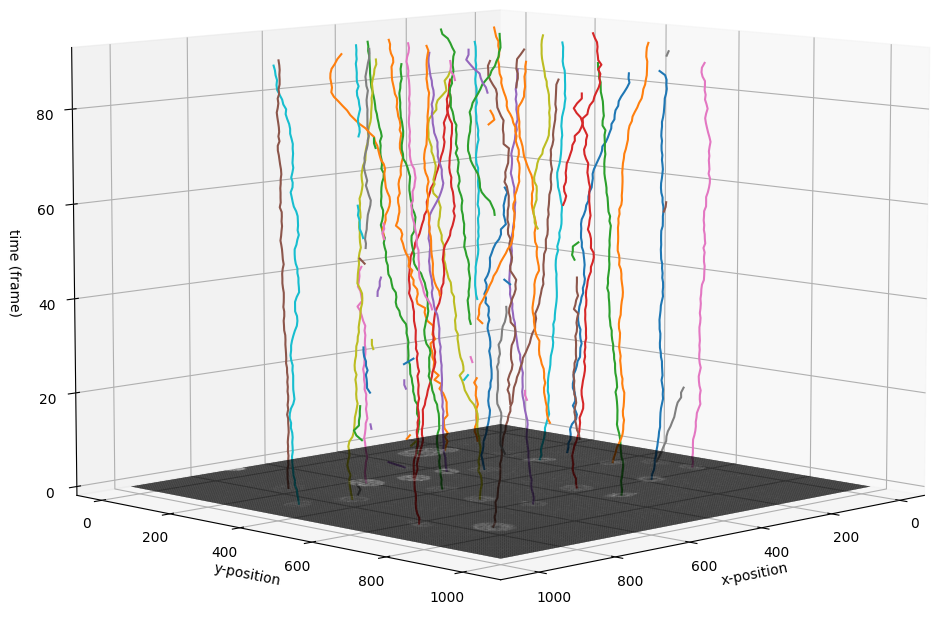}}\\
\subfloat[HeLa-01]{\includegraphics[width = 0.4\textwidth]{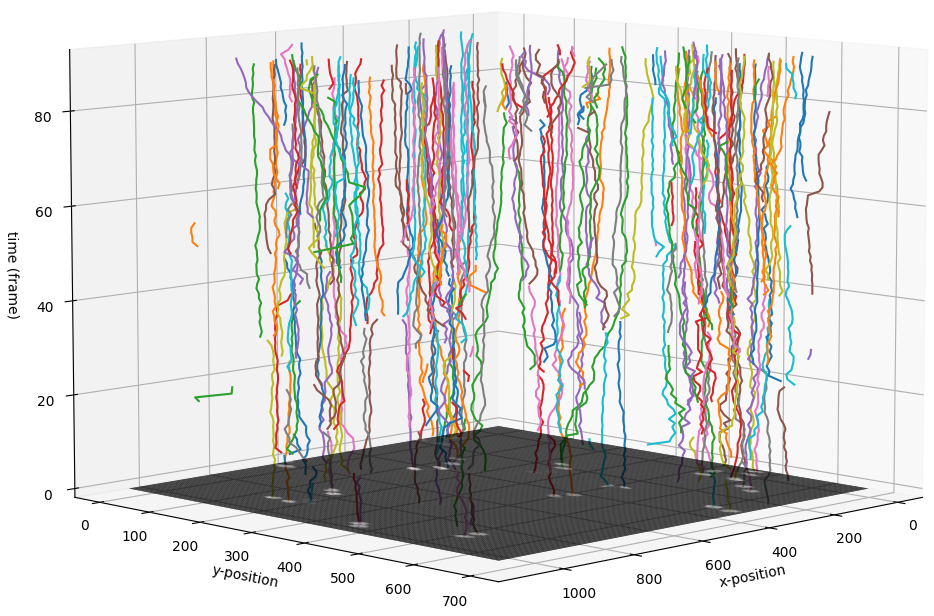}}~
\subfloat[HeLa-02]{\includegraphics[width = 0.4\textwidth]{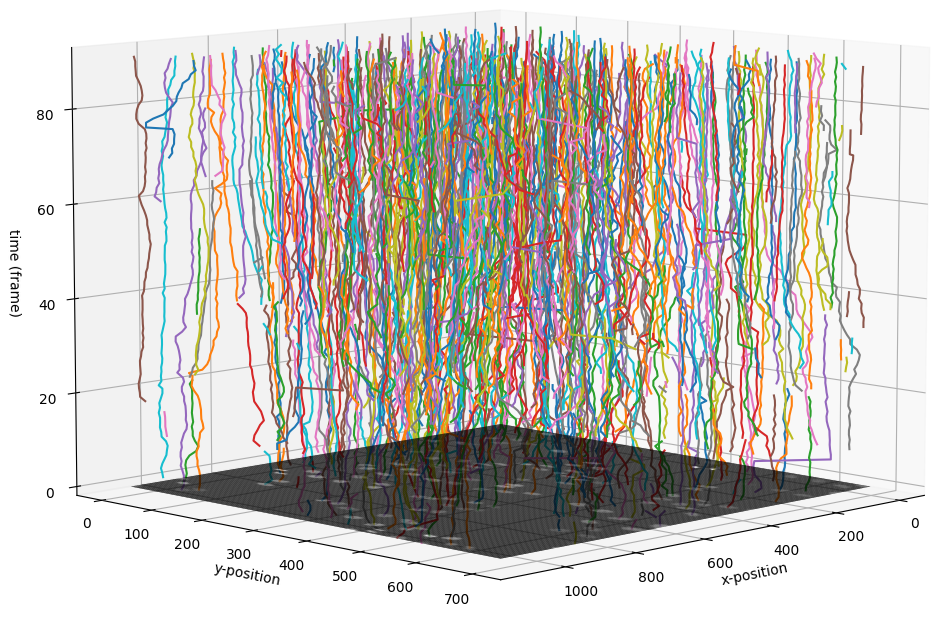}}\\
\subfloat[U373-01]{\includegraphics[width = 0.4\textwidth]{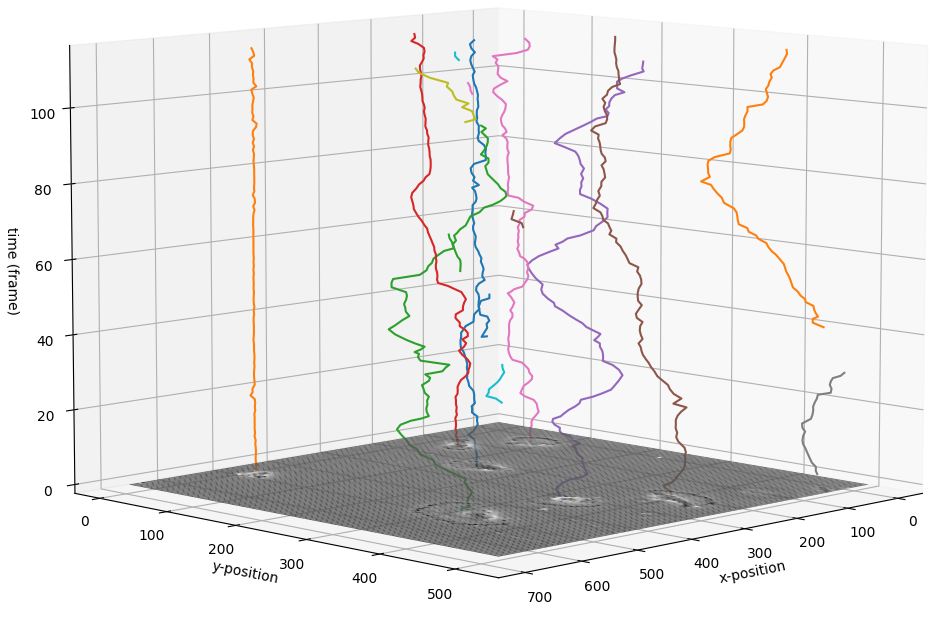}}~
\subfloat[U373-02]{\includegraphics[width = 0.4\textwidth]{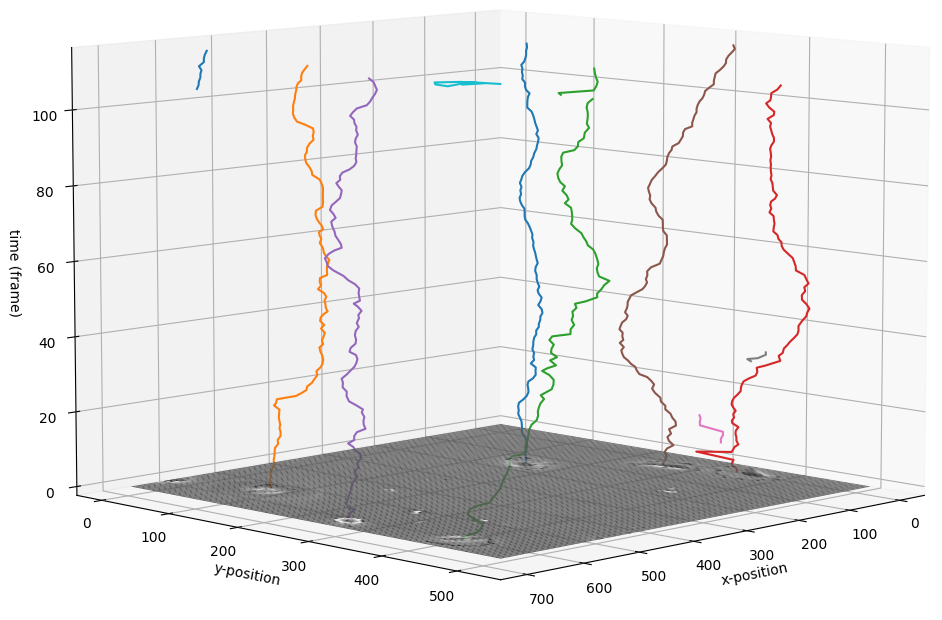}}\\
\caption[Visualization of the generated tracking trees in the CTC training datasets.]{Visualization of the generated tracking trees in the CTC~\cite{isbi} training datasets.}
\label{fig:tracking_trees}
\end{figure}

Finally, we provide visual detection results on the CTC~\cite{isbi} datasets on Figure~\ref{fig:visual_results}. We can observe that EBBs provide a good description of the cell shapes for the GOWT1 and HeLa datasets, but not so much for the U373 datasets. We can also note the high recall rate of the object detector, since almost all cells are retrieved. Figure~\ref{fig:tracking_trees} presents the generated tracking trees\footnote{We also provide animated images in our GitHub repository at: \url{https://github.com/LucasKirsten/Deep-Cell-Tracking-EBB/}}. Analyzing the results for the GOWT1 and U373 datasets, which are less cluttered, we can observe that our method could produce clear paths for most of the initial cells. These datasets have almost no mitosis or apoptosis events, so the paths that seem to emerge in later frames can be false positives or cells emerging from the image borders.

\subsection{Sensitivity analysis}\label{sec:abblation}

In this section, we analyze the sensitivity of our tracking-by-detection method regarding its hyper-parameters. We randomly sampled the parameter values within an interval and evaluated the results on the CTC~\cite{isbi} tested training datasets (i.e., with provided GT). The parameters were randomly sampled in the following scheme: $\lambda_{link}$ and $\lambda_{mit} \in \mathbb{R^+}$ linearly spaced between 5 and 1000 with step 25, $t_{th} \in \mathbb{N}$ linearly ranging from 1 to 8 with step 1, $\tau_s \in \mathbb{R^+}$ linearly ranging from 0.4 to 0.9 with 0.05 step, and $\tau_{FP} \in \mathbb{R^+}$ linearly ranging from 0.5 to 0.9 with 0.05 step. For consistently evaluating the impact of the parameters on the different datasets, we subtracted the metrics values from the ones reported in Table~\ref{tab:ISBI_results} when using the watershed method to show the relative gain or loss when changing the hyper-parameters.

\begin{figure}[]
\centering
\includegraphics[width = 0.48\textwidth]{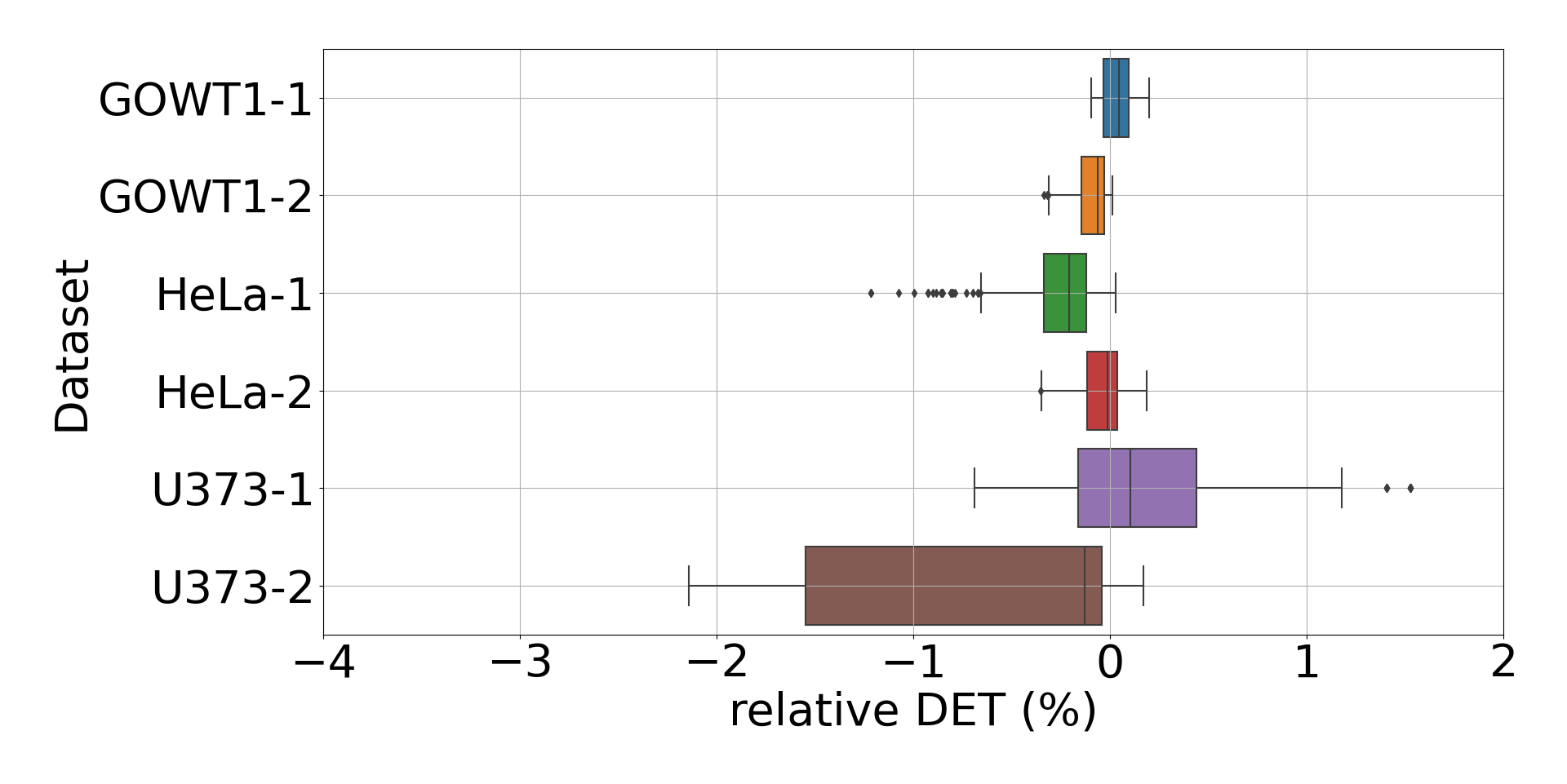} ~
\includegraphics[width = 0.48\textwidth]{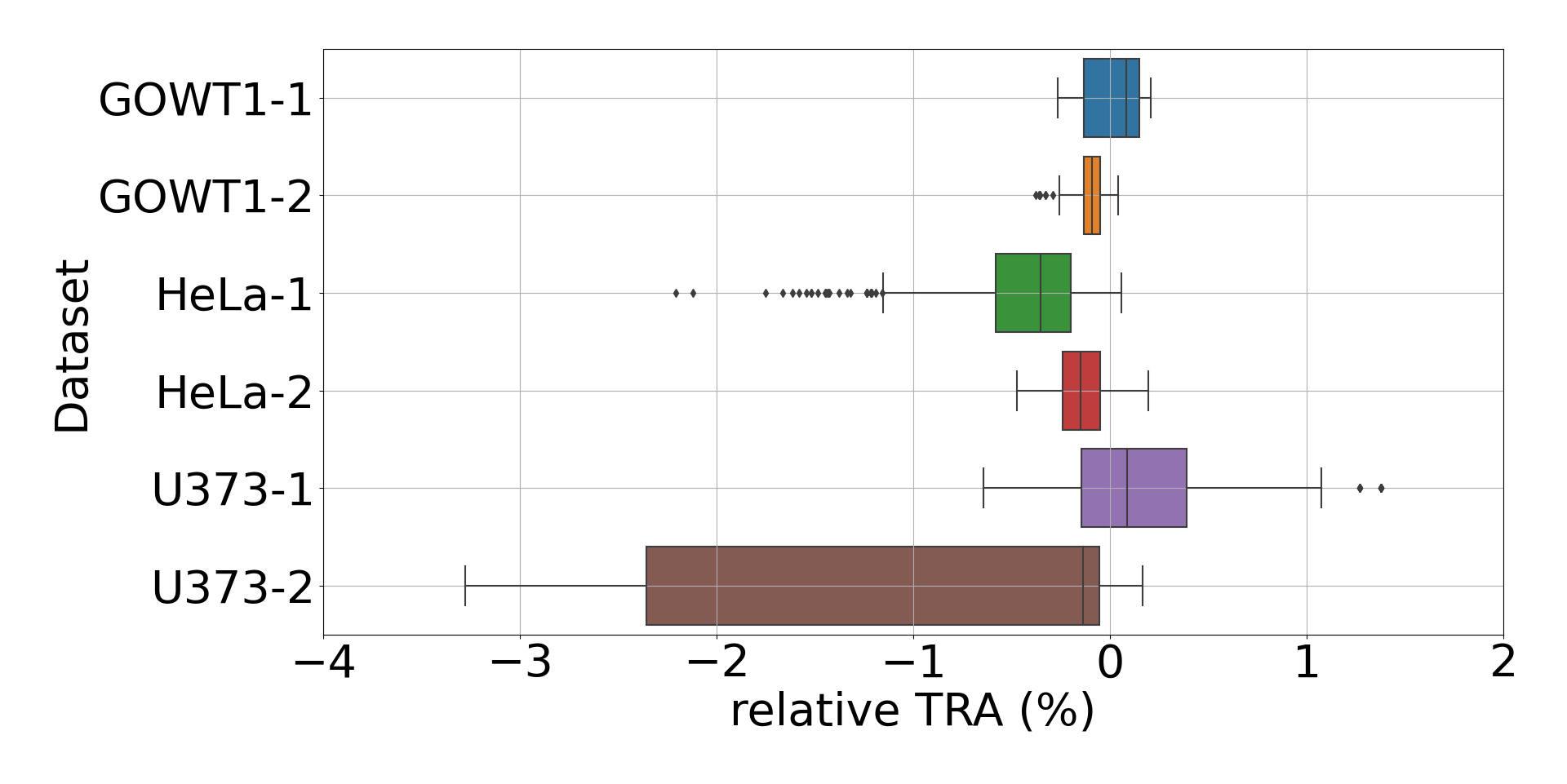}\\
\caption[Impacts on the individual datasets evaluation from randomly sampling the hyper-parameters.]{Impacts on the individual datasets evaluation from randomly sampling the hyper-parameters. The zero value refers to an output metric value equal to the one reported in Table~\ref{tab:ISBI_results}.}
\label{fig:abblation}
\end{figure}

Figure~\ref{fig:abblation} shows a boxplot with the relative changes of the DET and TRA metrics for a set of $\sim\!400$ random combinations of the hyper-parameters in each individual dataset. We can observe that the impact of changing the parameters is small for most of the datasets. The worst-case scenario occurs on the U373-2 dataset, with a negative impact of $\sim\!3.5\%$ on the TRA metric. On the other hand, we note that some combination of hyper-parameters can actually improve the results obtained with the default parameters.

\begin{figure}[]
\centering
\subfloat[DET metric]{\includegraphics[width=.48\textwidth]{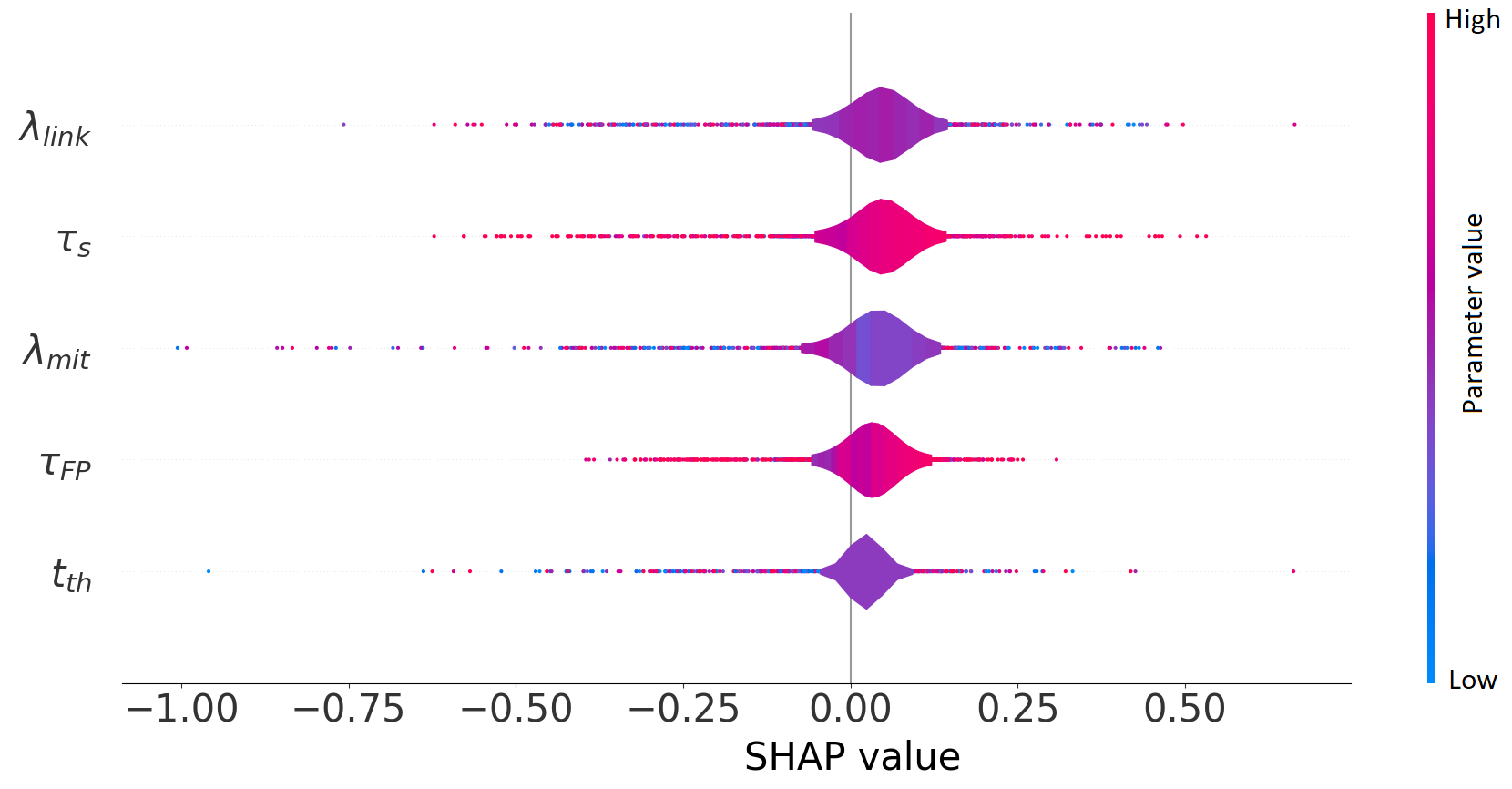}} ~
\subfloat[TRA metric]{\includegraphics[width=.48\textwidth]{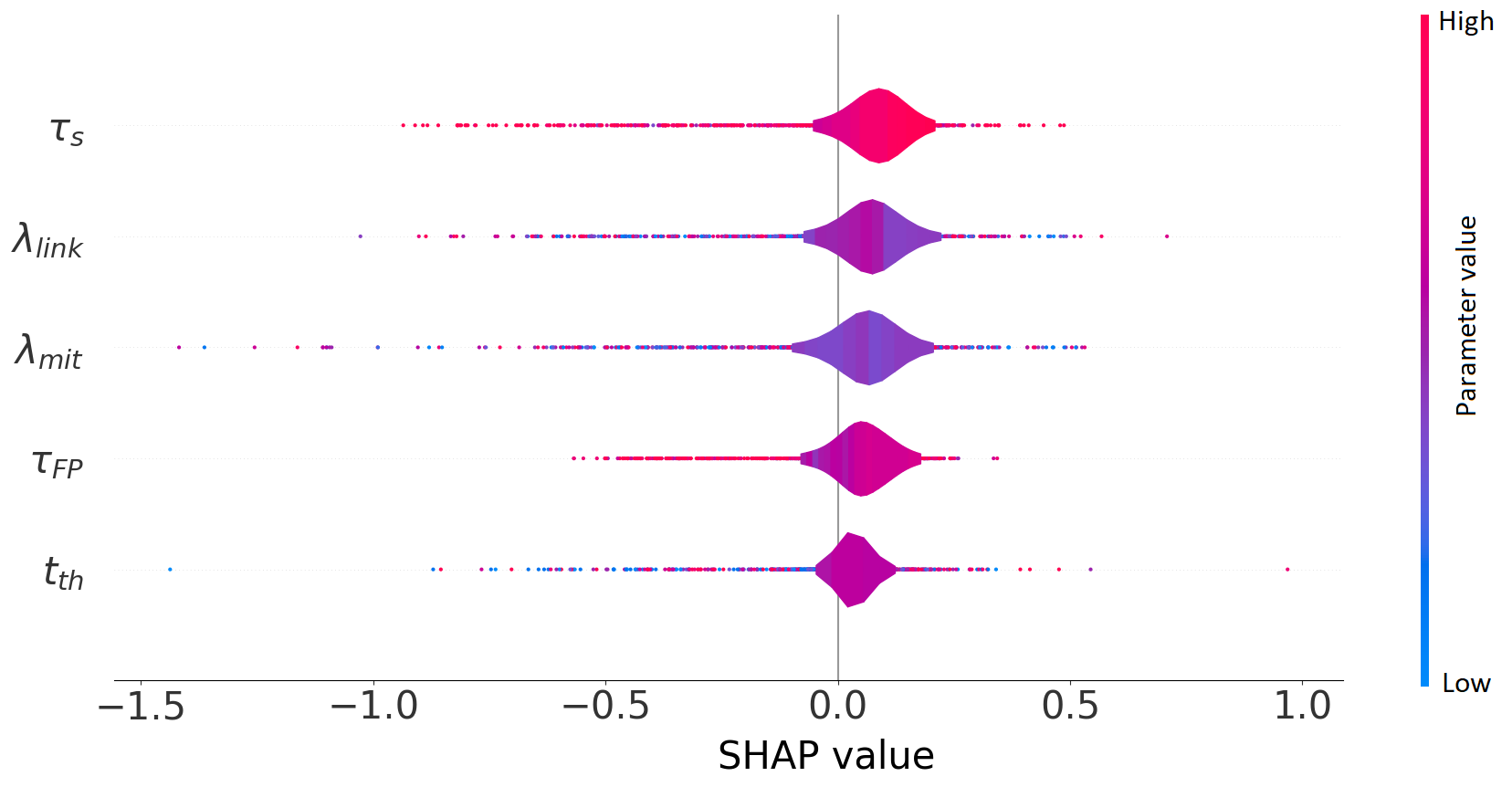}}
\caption[Individual impacts of the hyper-parameters on the method performance using the SHAP values regarding the DET and TRA metrics.]{Individual impacts of the hyper-parameters on the method performance using the SHAP values~\cite{shap,shap_plot} regarding the DET and TRA metrics.}
\label{fig:SHAP}
\end{figure}

In order to access the impact of the individual hyper-parameters on the method, we used the Shapley Additive Explanations (SHAP values)~\cite{shap,shap_plot} using the DET and TRA evaluation metrics as the targets and averaging the results among all the CTC tested datasets. SHAP values provide insights into feature contributions, distributing credit among features to explain machine learning predictions. They measure the impact of each feature compared to its absence or average value, allowing a nuanced understanding of feature importance. Positive values increase predictions, negative values decrease predictions, and zero values have no impact. Examining SHAP values helps identify influential features, aiding feature selection, model debugging, and understanding model decisions. In our case, we adapted the method to work with the hyper-parameters instead of the features. Figure~\ref{fig:SHAP} presents the violin plots, where the color indicates the parameter value and the horizontal axis denotes the corresponding SHAP value. The small range of the horizontal axis and concentration at small SHAP values indicate that the method is robust to the parameter choice. Furthermore, we note that most random combinations of individual parameters lead to a positive impact on the metric scores. 

\section{Conclusions}

We proposed a tracking-by-detection method that explores an OBB detector to identify cells,  represent them as ellipses, and then uses the detection information in an unsupervised tracking algorithm based on tracklet association. Our method alleviates the annotation efforts by representing the cells as a 5-parameter oriented ellipses (that can be either annotated as an OBB or EBB), and by defining an unsupervised tracking system oriented solely on the detection information retrieved by the trained object detector. Our results demonstrate that the cell elliptical representation presents a good approximation for the full segmentation mask, particularly for lineages with a regular shape. Furthermore, our tracking-by-detection method can achieve results competitive to other state-of-the-art methods that require considerably more annotated data. Moreover, our method reduces the hardware requirements for training and predicting when compared to the current trend of end-to-end trackers, since it requires training only one object detector and does not rely on training a complete detection and association deep learning architecture that needs both batches of frame images and their objects associations. We believe that our method can be broadly used in applications where there are limited resources or short deadlines for retrieving the full annotations.

\bibliographystyle{ieeetr}
\bibliography{references}

\begin{thebibliography}{10}

\bibitem{syed2008detection}
T.~Q. Syed, V.~Vigneron, S.~Lelandais, G.~Barlovatz-Meimon, M.~Malo, C.~Charri{\`e}re-Bertrand, and C.~Montagne, ``Detection and counting of" in vivo" cells to predict cell migratory potential,'' in {\em 2008 First Workshops on Image Processing Theory, Tools and Applications}, pp.~1--8, IEEE, 2008.

\bibitem{leite2015computational}
M.~R.~C. Leite, I.~A. Cestari, and I.~N. Cestari, ``Computational tool for morphological analysis of cultured neonatal rat cardiomyocytes,'' in {\em 2015 37th Annual International Conference of the IEEE Engineering in Medicine and Biology Society (EMBC)}, pp.~3517--3520, IEEE, 2015.

\bibitem{di2019learning}
D.~Di~Giuseppe, F.~Corsi, A.~Mencattini, M.~C. Comes, P.~Casti, C.~Di~Natale, L.~Ghibelli, and E.~Martinelli, ``Learning cancer-related drug efficacy exploiting consensus in coordinated motility within cell clusters,'' {\em IEEE Transactions on Biomedical Engineering}, vol.~66, no.~10, pp.~2882--2888, 2019.

\bibitem{gradeci2020single}
D.~Gradeci, A.~Bove, G.~Charras, A.~R. Lowe, and S.~Banerjee, ``Single-cell approaches to cell competition: high-throughput imaging, machine learning and simulations,'' in {\em Seminars in cancer biology}, vol.~63, pp.~60--68, Elsevier, 2020.

\bibitem{hayashida2022consistent}
J.~Hayashida, K.~Nishimura, and R.~Bise, ``Consistent cell tracking in multi-frames with spatio-temporal context by object-level warping loss,'' in {\em Proceedings of the IEEE/CVF Winter Conference on Applications of Computer Vision}, pp.~1727--1736, 2022.

\bibitem{emami2021computerized}
N.~Emami, Z.~Sedaei, and R.~Ferdousi, ``Computerized cell tracking: Current methods, tools and challenges,'' {\em Visual Informatics}, vol.~5, no.~1, pp.~1--13, 2021.

\bibitem{ulman2017objective}
V.~Ulman, M.~Ma{\v{s}}ka, K.~E. Magnusson, O.~Ronneberger, C.~Haubold, N.~Harder, P.~Matula, P.~Matula, D.~Svoboda, M.~Radojevic, {\em et~al.}, ``An objective comparison of cell-tracking algorithms,'' {\em Nature methods}, vol.~14, no.~12, pp.~1141--1152, 2017.

\bibitem{cpn}
S.~U. Akram, J.~Kannala, L.~Eklund, and J.~Heikkil{\"a}, ``Cell tracking via proposal generation and selection,'' {\em arXiv preprint arXiv:1705.03386}, 2017.

\bibitem{maskrcnn}
K.~He, G.~Gkioxari, P.~Doll{\'a}r, and R.~Girshick, ``Mask r-cnn,'' in {\em Proceedings of the IEEE international conference on computer vision}, pp.~2961--2969, 2017.

\bibitem{bise}
R.~Bise, Z.~Yin, and T.~Kanade, ``Reliable cell tracking by global data association,'' in {\em 2011 IEEE International Symposium on Biomedical Imaging: From Nano to Macro}, pp.~1004--1010, IEEE, 2011.

\bibitem{unet}
O.~Ronneberger, P.~Fischer, and T.~Brox, ``U-net: Convolutional networks for biomedical image segmentation,'' in {\em International Conference on Medical image computing and computer-assisted intervention}, pp.~234--241, Springer, 2015.

\bibitem{gcme}
R.~Bensch and O.~Ronneberger, ``Cell segmentation and tracking in phase contrast images using graph cut with asymmetric boundary costs,'' in {\em 2015 IEEE 12th International Symposium on Biomedical Imaging (ISBI)}, pp.~1220--1223, IEEE, 2015.

\bibitem{unets}
D.~K. Gupta, N.~de~Bruijn, A.~Panteli, and E.~Gavves, ``Tracking-assisted segmentation of biological cells,'' {\em arXiv preprint arXiv:1910.08735}, 2019.

\bibitem{drl}
J.~Wang, X.~Su, L.~Zhao, and J.~Zhang, ``Deep reinforcement learning for data association in cell tracking,'' {\em Frontiers in Bioengineering and Biotechnology}, vol.~8, p.~298, 2020.

\bibitem{kulharia2020box2seg}
V.~Kulharia, S.~Chandra, A.~Agrawal, P.~Torr, and A.~Tyagi, ``Box2seg: Attention weighted loss and discriminative feature learning for weakly supervised segmentation,'' in {\em European Conference on Computer Vision}, pp.~290--308, Springer, 2020.

\bibitem{xing2017deep}
F.~Xing, Y.~Xie, H.~Su, F.~Liu, and L.~Yang, ``Deep learning in microscopy image analysis: A survey,'' {\em IEEE Transactions on Neural Networks and Learning Systems}, vol.~29, no.~10, pp.~4550--4568, 2017.

\bibitem{moen2019deep}
E.~Moen, D.~Bannon, T.~Kudo, W.~Graf, M.~Covert, and D.~Van~Valen, ``Deep learning for cellular image analysis,'' {\em Nature methods}, vol.~16, no.~12, pp.~1233--1246, 2019.

\bibitem{schmidt2018cell}
U.~Schmidt, M.~Weigert, C.~Broaddus, and G.~Myers, ``Cell detection with star-convex polygons,'' in {\em International Conference on Medical Image Computing and Computer-Assisted Intervention}, pp.~265--273, Springer, 2018.

\bibitem{newell2017associative}
A.~Newell, Z.~Huang, and J.~Deng, ``Associative embedding: End-to-end learning for joint detection and grouping,'' in {\em Advances in neural information processing systems}, pp.~2277--2287, 2017.

\bibitem{payer2018instance}
C.~Payer, D.~{\v{S}}tern, T.~Neff, H.~Bischof, and M.~Urschler, ``Instance segmentation and tracking with cosine embeddings and recurrent hourglass networks,'' in {\em International Conference on Medical Image Computing and Computer-Assisted Intervention}, pp.~3--11, Springer, 2018.

\bibitem{payer2019segmenting}
C.~Payer, D.~{\v{S}}tern, M.~Feiner, H.~Bischof, and M.~Urschler, ``Segmenting and tracking cell instances with cosine embeddings and recurrent hourglass networks,'' {\em Medical image analysis}, vol.~57, pp.~106--119, 2019.

\bibitem{zhao2021faster}
M.~Zhao, A.~Jha, Q.~Liu, B.~A. Millis, A.~Mahadevan-Jansen, L.~Lu, B.~A. Landman, M.~J. Tyska, and Y.~Huo, ``Faster mean-shift: Gpu-accelerated clustering for cosine embedding-based cell segmentation and tracking,'' {\em Medical Image Analysis}, vol.~71, p.~102048, 2021.

\bibitem{liu2021panoptic}
D.~Liu, D.~Zhang, Y.~Song, H.~Huang, and W.~Cai, ``Panoptic feature fusion net: a novel instance segmentation paradigm for biomedical and biological images,'' {\em IEEE Transactions on Image Processing}, vol.~30, pp.~2045--2059, 2021.

\bibitem{zhao2018deep}
Z.~Zhao, L.~Yang, H.~Zheng, I.~H. Guldner, S.~Zhang, and D.~Z. Chen, ``Deep learning based instance segmentation in 3d biomedical images using weak annotation,'' in {\em International Conference on Medical Image Computing and Computer-Assisted Intervention}, pp.~352--360, Springer, 2018.

\bibitem{zhao2020weakly}
T.~Zhao and Z.~Yin, ``Weakly supervised cell segmentation by point annotation,'' {\em IEEE Transactions on Medical Imaging}, vol.~40, no.~10, pp.~2736--2747, 2020.

\bibitem{oh2022scribble}
H.-J. Oh, K.~Lee, and W.-K. Jeong, ``Scribble-supervised cell segmentation using multiscale contrastive regularization,'' in {\em 2022 IEEE 19th International Symposium on Biomedical Imaging (ISBI)}, pp.~1--5, IEEE, 2022.

\bibitem{lu2021biofabrication}
K.~Lu, Y.~Qian, J.~Gong, Z.~Zhu, J.~Yin, L.~Ma, M.~Yu, and H.~Wang, ``Biofabrication of aligned structures that guide cell orientation and applications in tissue engineering,'' {\em Bio-Design and Manufacturing}, vol.~4, no.~2, pp.~258--277, 2021.

\bibitem{liu2016ssd}
W.~Liu, D.~Anguelov, D.~Erhan, C.~Szegedy, S.~Reed, C.-Y. Fu, and A.~C. Berg, ``Ssd: Single shot multibox detector,'' in {\em European conference on computer vision}, pp.~21--37, Springer, 2016.

\bibitem{redmon2016you}
J.~Redmon, S.~Divvala, R.~Girshick, and A.~Farhadi, ``You only look once: Unified, real-time object detection,'' in {\em Proceedings of the IEEE conference on computer vision and pattern recognition}, pp.~779--788, 2016.

\bibitem{carion2020end}
N.~Carion, F.~Massa, G.~Synnaeve, N.~Usunier, A.~Kirillov, and S.~Zagoruyko, ``End-to-end object detection with transformers,'' in {\em European conference on computer vision}, pp.~213--229, Springer, 2020.

\bibitem{tan2020efficientdet}
M.~Tan, R.~Pang, and Q.~V. Le, ``Efficientdet: Scalable and efficient object detection,'' in {\em Proceedings of the IEEE/CVF conference on computer vision and pattern recognition}, pp.~10781--10790, 2020.

\bibitem{Wang_2021_CVPR}
C.-Y. Wang, A.~Bochkovskiy, and H.-Y.~M. Liao, ``Scaled-yolov4: Scaling cross stage partial network,'' in {\em Proceedings of the IEEE/CVF Conference on Computer Vision and Pattern Recognition (CVPR)}, pp.~13029--13038, June 2021.

\bibitem{yang2022focal}
J.~Yang, C.~Li, and J.~Gao, ``Focal modulation networks,'' {\em arXiv preprint arXiv:2203.11926}, 2022.

\bibitem{mandal2021splinedist}
S.~Mandal and V.~Uhlmann, ``Splinedist: Automated cell segmentation with spline curves,'' in {\em 2021 IEEE 18th International Symposium on Biomedical Imaging (ISBI)}, pp.~1082--1086, IEEE, 2021.

\bibitem{r2cnn}
Y.~Jiang, X.~Zhu, X.~Wang, S.~Yang, W.~Li, H.~Wang, P.~Fu, and Z.~Luo, ``R2cnn: Rotational region cnn for orientation robust scene text detection,'' {\em arXiv preprint arXiv:1706.09579}, 2017.

\bibitem{yang2021r3det}
X.~Yang, J.~Yan, Z.~Feng, and T.~He, ``R3det: Refined single-stage detector with feature refinement for rotating object,'' in {\em AAAI}, 2021.

\bibitem{dota}
G.-S. Xia, X.~Bai, J.~Ding, Z.~Zhu, S.~Belongie, J.~Luo, M.~Datcu, M.~Pelillo, and L.~Zhang, ``Dota: A large-scale dataset for object detection in aerial images,'' in {\em The IEEE Conference on Computer Vision and Pattern Recognition (CVPR)}, June 2018.

\bibitem{kld}
X.~Yang, X.~Yang, J.~Yang, Q.~Ming, W.~Wang, Q.~Tian, and J.~Yan, ``Learning high-precision bounding box for rotated object detection via kullback-leibler divergence,'' in {\em Advances in Neural Information Processing Systems} (M.~Ranzato, A.~Beygelzimer, Y.~Dauphin, P.~Liang, and J.~W. Vaughan, eds.), vol.~34, pp.~18381--18394, Curran Associates, Inc., 2021.

\bibitem{isbi}
M.~Ma{\v{s}}ka, V.~Ulman, D.~Svoboda, P.~Matula, P.~Matula, C.~Ederra, A.~Urbiola, T.~Espa{\~n}a, S.~Venkatesan, D.~M. Balak, {\em et~al.}, ``A benchmark for comparison of cell tracking algorithms,'' {\em Bioinformatics}, vol.~30, no.~11, pp.~1609--1617, 2014.

\bibitem{epflheid}
E.~T{\"u}retken, X.~Wang, C.~Becker, C.~Haubold, and P.~Fua, ``Globally optimal cell tracking using integer programming,'' {\em arXiv preprint arXiv:1501.05499}, 2015.

\bibitem{blob}
S.~U. Akram, J.~Kannala, L.~Eklund, and J.~Heikkil{\"a}, ``Joint cell segmentation and tracking using cell proposals,'' in {\em 2016 IEEE 13th International Symposium on Biomedical Imaging (ISBI)}, pp.~920--924, IEEE, 2016.

\bibitem{nishimura2020weakly}
K.~Nishimura, J.~Hayashida, C.~Wang, D.~F.~E. Ker, and R.~Bise, ``Weakly-supervised cell tracking via backward-and-forward propagation,'' in {\em European Conference on Computer Vision}, pp.~104--121, Springer, 2020.

\bibitem{kth}
K.~E. Magnusson and J.~Jald{\'e}n, ``A batch algorithm using iterative application of the viterbi algorithm to track cells and construct cell lineages,'' in {\em 2012 9th IEEE International Symposium on Biomedical Imaging (ISBI)}, pp.~382--385, IEEE, 2012.

\bibitem{boukari2018automated}
F.~Boukari and S.~Makrogiannis, ``Automated cell tracking using motion prediction-based matching and event handling,'' {\em IEEE/ACM transactions on computational biology and bioinformatics}, vol.~17, no.~3, pp.~959--971, 2018.

\bibitem{xu2019automated}
B.~Xu, J.~Shi, M.~Lu, J.~Cong, L.~Wang, and B.~Nener, ``An automated cell tracking approach with multi-bernoulli filtering and ant colony labor division,'' {\em IEEE/ACM transactions on computational biology and bioinformatics}, vol.~18, no.~5, pp.~1850--1863, 2019.

\bibitem{hirose2017spf}
O.~Hirose, S.~Kawaguchi, T.~Tokunaga, Y.~Toyoshima, T.~Teramoto, S.~Kuge, T.~Ishihara, Y.~Iino, and R.~Yoshida, ``Spf-celltracker: Tracking multiple cells with strongly-correlated moves using a spatial particle filter,'' {\em IEEE/ACM transactions on computational biology and bioinformatics}, vol.~15, no.~6, pp.~1822--1831, 2017.

\bibitem{probiou}
J.~M. Llerena, L.~F. Zeni, L.~N. Kristen, and C.~Jung, ``Gaussian bounding boxes and probabilistic intersection-over-union for object detection,'' {\em arXiv preprint arXiv:2106.06072}, 2021.

\bibitem{chen2020piou}
Z.~Chen, K.~Chen, W.~Lin, J.~See, H.~Yu, Y.~Ke, and C.~Yang, ``Piou loss: Towards accurate oriented object detection in complex environments,'' in {\em European conference on computer vision}, pp.~195--211, Springer, 2020.

\bibitem{Murrugarra-Llerena_2022_CVPR}
J.~Murrugarra-Llerena, L.~N. Kirsten, and C.~R. Jung, ``Can we trust bounding box annotations for object detection?,'' in {\em Proceedings of the IEEE/CVF Conference on Computer Vision and Pattern Recognition (CVPR) Workshops}, pp.~4813--4822, June 2022.

\bibitem{gwd}
X.~Yang, J.~Yan, M.~Qi, W.~Wang, Z.~Xiaopeng, and T.~Qi, ``Rethinking rotated object detection with gaussian wasserstein distance loss,'' in {\em International Conference on Machine Learning (ICML)}, 2021.

\bibitem{bhattacharyya1946measure}
A.~Bhattacharyya, ``On a measure of divergence between two multinomial populations,'' {\em Sankhy{\=a}: the indian journal of statistics}, pp.~401--406, 1946.

\bibitem{hellinger1909neue}
E.~Hellinger, ``Neue begr{\"u}ndung der theorie quadratischer formen von unendlichvielen ver{\"a}nderlichen.,'' {\em Journal f{\"u}r die reine und angewandte Mathematik}, vol.~1909, no.~136, pp.~210--271, 1909.

\bibitem{kailath1967divergence}
T.~Kailath, ``The divergence and bhattacharyya distance measures in signal selection,'' {\em IEEE transactions on communication technology}, vol.~15, no.~1, pp.~52--60, 1967.

\bibitem{kuhn1955hungarian}
H.~W. Kuhn, ``The hungarian method for the assignment problem,'' {\em Naval research logistics quarterly}, vol.~2, no.~1-2, pp.~83--97, 1955.

\bibitem{huh2010automated}
S.~Huh, R.~Bise, M.~Chen, T.~Kanade, {\em et~al.}, ``Automated mitosis detection of stem cell populations in phase-contrast microscopy images,'' {\em IEEE transactions on medical imaging}, vol.~30, no.~3, pp.~586--596, 2010.

\bibitem{opencv}
G.~Bradski, ``{The OpenCV Library},'' {\em Dr. Dobb's Journal of Software Tools}, 2000.

\bibitem{liu2020deep}
L.~Liu, W.~Ouyang, X.~Wang, P.~Fieguth, J.~Chen, X.~Liu, and M.~Pietik{\"a}inen, ``Deep learning for generic object detection: A survey,'' {\em International journal of computer vision}, vol.~128, no.~2, pp.~261--318, 2020.

\bibitem{resnet}
K.~He, X.~Zhang, S.~Ren, and J.~Sun, ``Deep residual learning for image recognition,'' in {\em Proceedings of the IEEE conference on computer vision and pattern recognition}, pp.~770--778, 2016.

\bibitem{imagenet}
J.~Deng, W.~Dong, R.~Socher, L.-J. Li, K.~Li, and L.~Fei-Fei, ``Imagenet: A large-scale hierarchical image database,'' in {\em 2009 IEEE conference on computer vision and pattern recognition}, pp.~248--255, Ieee, 2009.

\bibitem{cbc}
J.~Forrest, T.~Ralphs, H.~G. Santos, S.~Vigerske, J.~Forrest, L.~Hafer, B.~Kristjansson, jpfasano, EdwinStraver, M.~Lubin, rlougee, jpgoncal1, Jan-Willem, h-i gassmann, S.~Brito, Cristina, M.~Saltzman, tosttost, B.~Pitrus, F.~MATSUSHIMA, and to~st, ``coin-or/cbc: Release releases/2.10.8,'' May 2022.

\bibitem{python}
G.~Van~Rossum and F.~L. Drake~Jr, {\em Python reference manual}.
\newblock Centrum voor Wiskunde en Informatica Amsterdam, 1995.

\bibitem{mota}
K.~Bernardin and R.~Stiefelhagen, ``Evaluating multiple object tracking performance: the clear mot metrics,'' {\em EURASIP Journal on Image and Video Processing}, vol.~2008, pp.~1--10, 2008.

\bibitem{mot16}
A.~Milan, L.~Leal-Taix{\'e}, I.~Reid, S.~Roth, and K.~Schindler, ``Mot16: A benchmark for multi-object tracking,'' {\em arXiv preprint arXiv:1603.00831}, 2016.

\bibitem{mota_id}
E.~Ristani, F.~Solera, R.~Zou, R.~Cucchiara, and C.~Tomasi, ``Performance measures and a data set for multi-target, multi-camera tracking,'' in {\em European conference on computer vision}, pp.~17--35, Springer, 2016.

\bibitem{boukari2016joint}
F.~Boukari and S.~Makrogiannis, ``Joint level-set and spatio-temporal motion detection for cell segmentation,'' {\em BMC Medical Genomics}, vol.~9, no.~2, pp.~179--194, 2016.

\bibitem{shap}
S.~M. Lundberg and S.-I. Lee, ``A unified approach to interpreting model predictions,'' in {\em Advances in Neural Information Processing Systems 30} (I.~Guyon, U.~V. Luxburg, S.~Bengio, H.~Wallach, R.~Fergus, S.~Vishwanathan, and R.~Garnett, eds.), pp.~4765--4774, Curran Associates, Inc., 2017.

\bibitem{shap_plot}
S.~M. Lundberg, B.~Nair, M.~S. Vavilala, M.~Horibe, M.~J. Eisses, T.~Adams, D.~E. Liston, D.~K.-W. Low, S.-F. Newman, J.~Kim, {\em et~al.}, ``Explainable machine-learning predictions for the prevention of hypoxaemia during surgery,'' {\em Nature Biomedical Engineering}, vol.~2, no.~10, p.~749, 2018.

\end{thebibliography}

\end{document}